\documentclass[a4paper,USenglish,cleveref,autoref,thm-restate]{oasics-v2021}

\usepackage{soul}
\usepackage{csquotes}
\usepackage{graphicx}
\usepackage{amsmath}
\usepackage{booktabs}
\usepackage{algorithm}
\usepackage{algorithmicx}
\usepackage[noEnd=true]{algpseudocodex}

\usepackage{acronym}
\usepackage[createShortEnv,conf={no link to proof, restate, text proof={Proof}}]{proof-at-the-end}
\usepackage[textsize=scriptsize,colorinlistoftodos,obeyDraft]{todonotes}
\usepackage{mfirstuc}
\usepackage{siunitx}
\usepackage[mode=buildnew]{standalone}

\newcommand*{\smallfont}{}

\newenvironment{sizeddisplay}
 {\par\nobreak\noindent\ignorespaces}
 {\ignorespacesafterend}

\usepackage{ifdraft}
\ifdraft{\pagestyle{plain}}{}

\nolinenumbers

\newboolean{showappendix}
\setboolean{showappendix}{true}

\newif\ifextendedappendix
\extendedappendixtrue

\newif\ifproofideas
\proofideastrue 

\usepackage{comment}
\ifproofideas
  \specialcomment{proofidea}{\begin{proof}[Proof Idea]}{\end{proof}}
\else
  \excludecomment{proofidea}
\fi

\usepackage{xspace}
\usepackage{mfirstuc}
\usepackage{mathtools}
\usepackage{amsfonts}

\newcommand*{\talocs}{\ensuremath{L}}
\newcommand*{\taloc}{\ensuremath{l}}
\newcommand*{\taalph}{\ensuremath{\Sigma}}
\newcommand*{\taclocks}{\ensuremath{X}}
\newcommand*{\taswitches}{\ensuremath{E}}
\newcommand*{\talts}{\ensuremath{\mathcal{S}_{\ta}}\xspace}
\newcommand*{\taltsstates}{\ensuremath{Q_\ta}}
\newcommand*{\taltsstate}{\ensuremath{q_\ta}}
\newcommand*{\taltsalph}{\ensuremath{\Sigma_\ta}}
\newcommand*{\taltstrans}[2]{\ensuremath{\xrightarrow[#1]{#2}}\xspace}
\newcommand*{\clockvaluation}{\clockval}
\newcommand*{\clockconstraint}{\ensuremath{g}\xspace}
\newcommand*{\smid}{\ensuremath{\:\mid\:}}
\newcommand*{\clockconstraints}{\ensuremath{\Phi}\xspace}
\DeclareMathOperator{\runs}{Runs}

\newcommand*{\finruns}{\runs^*}

\newcommand*{\accfinruns}{\runs^*_F}

\newcommand*{\lang}{\ensuremath{\mathcal{L}}\xspace}

\newcommand*{\tastate}{\ensuremath{l}\xspace}
\DeclareMathOperator{\tw}{tw}

\newcommand*{\ltsstate}{\taltsstate}
\newcommand*{\ltstrans}[1][]{\ensuremath{\xrightarrow{#1}}}

\newcommand*{\eqdef}{\ensuremath{\coloneqq}}
\newcommand*{\equivspace}{\;\equiv\;}
\newcommand*{\la}{\langle}
\newcommand*{\ra}{\rangle}
\newcommand*{\reals}{\ensuremath{\mathbb{R}}\xspace}
\newcommand*{\realpos}{\ensuremath{\mathbb{R}_{\geq 0}}\xspace}
\newcommand*{\naturals}{\ensuremath{\mathbb{N}}\xspace}
\newcommand*{\regop}{\ensuremath{\mathcal{R}}\xspace}
\newcommand*{\reg}[1]{\ensuremath{\regop[#1]}\xspace}

\newcommand*{\golog}{\textsc{Golog}\xspace}
\newcommand*{\congolog}{\textsc{ConGolog}\xspace}
\newcommand*{\indigolog}{\textsc{IndiGolog}\xspace}

\newcommand*{\situations}{\ensuremath{\mathcal{S}}\xspace}
\newcommand*{\actions}{\ensuremath{\mathcal{A}}\xspace}
\newcommand*{\objects}{\ensuremath{\mathcal{O}}\xspace}
\newcommand*{\fluents}{\ensuremath{\mathcal{F}}\xspace}
\newcommand*{\clocks}{\ensuremath{\mathcal{C}}\xspace}

\makeatletter
\@ifclassloaded{beamer}
{}
{
\newcommand*{\action}[1]{\ensuremath{\mathit{#1}}\xspace}
}
\makeatother
\newcommand*{\rfluent}[1]{\ensuremath{\xcapitalisewords{\mathit{#1}}}\xspace}
\newcommand*{\ffluent}[1]{\ensuremath{\mathit{#1}}\xspace}

\DeclareMathOperator*{\doopsym}{do}
\newcommand*{\doop}{\ensuremath{\doopsym}\xspace}
\newcommand*{\bat}{\ensuremath{\mathcal{D}}\xspace}
\DeclareMathOperator{\poss}{Poss}
\DeclareMathOperator{\exec}{Exec}
\DeclareMathOperator{\atime}{time}
\DeclareMathOperator{\sstart}{start}

\newcommand*{\clockval}{\ensuremath{\nu}\xspace}
\newcommand*{\sac}[1]{\ensuremath{\action{s\xcapitalisewords{#1}}}\xspace}
\newcommand*{\eac}[1]{\ensuremath{\action{e\xcapitalisewords{#1}}}\xspace}
\newcommand*{\wait}{\action{wait}\xspace}

\newcommand*{\hasCoffee}{\rfluent{isFull}}
\newcommand*{\isStrong}{\rfluent{strong}}
\newcommand*{\brewing}{\rfluent{brew}}
\newcommand*{\pouring}{\rfluent{pour}}

\newcommand*{\maxconst}{\ensuremath{\mathcal{K}}\xspace}
\newcommand*{\tabisim}{\ensuremath{\approx_{\bat\maxconst}}\xspace}
\newcommand*{\clockequiv}{\ensuremath{\sim_\maxconst}\xspace}
\newcommand*{\regequiv}{\ensuremath{\cong_\maxconst}\xspace}
\DeclareMathOperator*{\fract}{fract}
\newcommand*{\mi}[1]{\ensuremath{\mathit{#1}}\xspace}
\newcommand*{\ta}{\ensuremath{\mathcal{A}}\xspace}

\newcommand*{\absts}[1][\bat]{\ensuremath{\mathcal{T}_{#1}}\xspace}

\DeclareMathOperator*{\nilop}{nil}
\newcommand*{\nil}{\ensuremath{\nilop}\xspace}

\DeclareMathOperator*{\gtransop}{Trans}
\newcommand*{\gtrans}{\ensuremath{\gtransop}\xspace}
\DeclareMathOperator*{\gfinalop}{Final}
\newcommand*{\gfinal}{\ensuremath{\gfinalop}\xspace}
\newcommand*{\true}{\textsc{True}\xspace}
\newcommand*{\false}{\textsc{False}\xspace}
\newcommand*{\gwhile}{\ensuremath{\;\mathbf{while}\;}}
\newcommand*{\gdo}{\ensuremath{\;\mathbf{do}\;}}
\newcommand*{\gdone}{\ensuremath{\;\mathbf{done}\;}}
\newcommand*{\gif}{\ensuremath{\;\mathbf{if}\;}}

\newcommand*{\gelse}{\ensuremath{\;\mathbf{else}\;}}
\newcommand*{\gfi}{\ensuremath{\;\mathbf{fi}\;}}

\newcommand*{\pvec}[1]{\vec{#1}\mkern2mu\vphantom{#1}}

\definecolor{rwth-blue}{cmyk}{1,.5,0,0}
\definecolor{rwth-lblue}{cmyk}{0.75,0.38,0,0}
\definecolor{rwth-llblue}{cmyk}{0.45,0.14,0,0}
\definecolor{rwth-lllblue}{cmyk}{0.23,0.07,0,0}
\definecolor{rwth-llllblue}{cmyk}{0.09,0.03,0,0}

\definecolor{rwth-black}{cmyk}{0,0,0,1}
\colorlet{rwth-lblack}{rwth-black!75}
\colorlet{rwth-llblack}{rwth-black!50}
\colorlet{rwth-lllblack}{rwth-black!25}
\colorlet{rwth-llllblack}{rwth-black!10}

\definecolor{rwth-magenta}{cmyk}{0,1,.25,0}
\colorlet{rwth-lmagenta}{rwth-magenta!75}
\colorlet{rwth-llmagenta}{rwth-magenta!50}
\colorlet{rwth-lllmagenta}{rwth-magenta!25}
\colorlet{rwth-llllmagenta}{rwth-magenta!10}

\definecolor{rwth-yellow}{cmyk}{0,0,1,0}
\colorlet{rwth-lyellow}{rwth-yellow!75}
\colorlet{rwth-llyellow}{rwth-yellow!50}
\colorlet{rwth-lllyellow}{rwth-yellow!25}
\colorlet{rwth-llllyellow}{rwth-yellow!10}

\definecolor{rwth-petrol}{cmyk}{1,0.3,0.5,0.3}
\colorlet{rwth-lpetrol}{rwth-petrol!75}
\colorlet{rwth-llpetrol}{rwth-petrol!50}
\colorlet{rwth-lllpetrol}{rwth-petrol!25}
\colorlet{rwth-llllpetrol}{rwth-petrol!10}

\definecolor{rwth-turquoise}{cmyk}{1,0,0.4,0}
\colorlet{rwth-lturquoise}{rwth-turquoise!75}
\colorlet{rwth-llturquoise}{rwth-turquoise!50}
\colorlet{rwth-lllturquoise}{rwth-turquoise!25}
\colorlet{rwth-llllturquoise}{rwth-turquoise!10}

\definecolor{rwth-green}{cmyk}{.7,0,1,0}
\colorlet{rwth-lgreen}{rwth-green!75}
\colorlet{rwth-llgreen}{rwth-green!50}
\colorlet{rwth-lllgreen}{rwth-green!25}
\colorlet{rwth-llllgreen}{rwth-green!10}

\definecolor{rwth-grass}{cmyk}{.35,0,1,0}
\colorlet{rwth-lgrass}{rwth-grass!75}
\colorlet{rwth-llgrass}{rwth-grass!50}
\colorlet{rwth-lllgrass}{rwth-grass!25}
\colorlet{rwth-llllgrass}{rwth-grass!10}

\definecolor{rwth-orange}{cmyk}{0,.4,1,0}
\colorlet{rwth-lorange}{rwth-orange!75}
\colorlet{rwth-llorange}{rwth-orange!50}
\colorlet{rwth-lllorange}{rwth-orange!25}
\colorlet{rwth-llllorange}{rwth-orange!10}

\definecolor{rwth-red}{cmyk}{.15,1,1,0}
\colorlet{rwth-lred}{rwth-red!75}
\colorlet{rwth-llred}{rwth-red!50}
\colorlet{rwth-lllred}{rwth-red!25}
\colorlet{rwth-llllred}{rwth-red!10}

\definecolor{rwth-burgundy}{cmyk}{0.25,1,0.7,0.2}
\colorlet{rwth-lburgundy}{rwth-burgundy!75}
\colorlet{rwth-llburgundy}{rwth-burgundy!50}
\colorlet{rwth-lllburgundy}{rwth-burgundy!25}
\colorlet{rwth-llllburgundy}{rwth-burgundy!10}

\definecolor{rwth-violet}{cmyk}{0.7,1,0.35,0.15}
\colorlet{rwth-lviolet}{rwth-violet!75}
\colorlet{rwth-llviolet}{rwth-violet!50}
\colorlet{rwth-lllviolet}{rwth-violet!25}
\colorlet{rwth-llllviolet}{rwth-violet!10}

\definecolor{rwth-purple}{cmyk}{0.6,0.6,0,0}
\colorlet{rwth-lpurple}{rwth-purple!75}
\colorlet{rwth-llpurple}{rwth-purple!50}
\colorlet{rwth-lllpurple}{rwth-purple!25}
\colorlet{rwth-llllpurple}{rwth-purple!10}

\definecolor{rwth-cyan}{cmyk}{1,0,.4,0}\colorlet{rwth-lcyan}{rwth-cyan!50}\colorlet{rwth-llcyan}{rwth-cyan!25}
\definecolor{rwth-teal}{cmyk}{1,.3,.5,.3}\colorlet{rwth-lteal}{rwth-teal!50}\colorlet{rwth-llteal}{rwth-teal!25}
\definecolor{rwth-silver}{cmyk}{.39,.31,.32,.14}
\definecolor{rwth-gold}{cmyk}{.35,.46,.7,.35}

\acrodef{TA}{timed automaton}
\acrodefplural{TA}[TAs]{timed automata}
\acrodef{BAT}{basic action theory}
\acrodefplural{BAT}[BATs]{basic action theories}
\acrodef{LTS}{labeled transition system}
\acrodef{SSA}{successor state axiom}
\acrodefindefinite{SSA}{an}{a}
\acrodef{UNA}{unique name axiom}
\acrodef{SC}{situation calculus}
\acrodef{2CM}{2-counter machine}

\title{Decidable Reasoning About Time in Finite-Domain Situation Calculus Theories}

\author{Till Hofmann}{RWTH Aachen University, Germany \and University of Oxford, UK}{till.hofmann@cs.rwth-aachen.de}{https://orcid.org/0000-0002-8621-5939}{}
\author{Stefan Schupp}{TU Wien, Austria}{}{https://orcid.org/0000-0002-2055-7581}{}
\author{Gerhard Lakemeyer}{RWTH Aachen University, Germany}{}{https://orcid.org/0000-0002-7363-7593}{}

\authorrunning{T.\,Hofmann, S.\,Schupp, and G.\,Lakemeyer}
\Copyright{Till Hofmann, Stefan Schupp, and Gerhard Lakemeyer}
\relatedversion{}

\EventEditors{AndreA Orlandini and Sophie Pinchinat}
\EventNoEds{2}
\EventLongTitle{33rd International Symposium on Temporal Representation and Reasoning (TIME 2026)}
\EventShortTitle{TIME 2026}
\EventAcronym{TIME}
\EventYear{2026}
\EventDate{September 1--3, 2026}
\EventLocation{Cork, Ireland}
\EventLogo{}
\SeriesVolume{146}
\ArticleNo{11}
\funding{
This research was supported by the EU ICT-48 2020 project TAILOR (No 952215). Travel funding to present this paper was provided by the EU project euroFMX (No 101299128).
}
\relatedversiondetails{Official version}{https://doi.org/10.4230/OASIcs.TIME.2026.11}

\begin{document}

\maketitle

\begin{abstract}
Representing time is crucial for cyber-physical systems and has been studied extensively in the \acl*{SC}.
The most commonly used approach
represents time by adding a real-valued function $\atime(a)$ that attaches a time point to each action and consequently to each situation.
We show that in this approach, checking whether there is a reachable situation that satisfies a given formula is undecidable, even when the domain contains only finitely many objects.
We present an alternative approach based on well-established results from timed automata theory by introducing clocks as real-valued fluents with restricted successor state axioms and comparison operators.
With this restriction, we can show that the reachability problem for finite-domain basic action theories is decidable.
Finally, we apply our results to Golog program realization by presenting a decidable procedure for determining an action sequence that is a successful execution of a given program.
\end{abstract}

\section{Introduction}

Time is a ubiquitous concept in various applications, especially in cases where digital systems interact with the real world such as applications in robotics or generally in cyber-physical systems.
Early approaches model discrete time, which allows modeling and reasoning about system states at discrete points in time.
This approach gives rise to the problem of choosing an appropriate granularity (resolution) of time to facilitate useful reasoning.
The introduction of models which incorporate continuous and dense notions of time circumvents this problem and allows for more expressive modeling and analysis approaches.
Dense time has been under extensive investigation in the formal methods community, e.g., in the analysis of hybrid systems which started with the introduction of \acp{TA}~\cite{alur_theory_1994}.

The \ac{SC} is a well-known first-order formalism for reasoning about dynamically changing worlds, where each change in the world is caused by an \emph{action}.
Among others, it permits a representation of continuous time by attaching a time of occurrence to each action and hence also each situation~\cite{pinto_reasoning_1995,reiter_natural_1996}.
In this paper, we focus on \emph{reachability}, i.e., checking whether there is a reachable situation where some formula $\phi$ holds.
We show that the standard representation of time already makes the reachability problem undecidable, even when the domain contains only finitely many objects.
To recover decidability, we adapt the core techniques of timed automata theory to the situation calculus: we introduce clocks as real-valued fluents that may only be reset or advanced uniformly and compared against natural numbers, and we show that region abstraction, adapted to situations as a time-abstract bisimulation, yields a finite abstraction of the otherwise infinite tree of situations.
This transfer lets us decide reachability for clocked basic action theories over finite domains.
We then extend this result to \golog programs and investigate \emph{program realization}. Given a \golog program, we present a decidable method for determining a corresponding action sequence that results in a final program configuration.

Reachability and program realization in this restricted setting are closely related to recent work on synthesis for \golog programs.
Reactive synthesis has been studied for the propositional situation calculus \cite{degiacomoReactiveSynthesisGolog2026} and for first-order agent programs against uncontrollable environments with LTL\textsubscript{f} goals \cite{hofmannLTLfSynthesisFirstorder2025}.
Our realization procedure complements these: rather than synthesizing a strategy against an adversarial environment, we focus on the fully controllable setting, but extend the formalism with continuous time.

This paper is organized as follows: We discuss related work in \autoref{sec:related-work} and present the foundations of the \ac{SC} in \autoref{sec:foundations}.
We discuss reachability in \autoref{sec:reachability}, where we show undecidability for the existing approaches and then show a decidable approach in \autoref{sec:clocks-in-sitcalc}.
In \autoref{sec:program-realization}, we apply these results to program realization and conclude in \autoref{sec:conclusion}.

\section{Related Work}\label{sec:related-work}
Already McCarthy~\cite{mccarthy_situations_1963} proposed a fluent function $\mi{time}$ that gives the value of time in a situation.
Elaborating on this idea, Gelfond et al.~\cite{gelfond_what_1991} described an extension where time is a fluent with values from the integers or real numbers and where each action has a duration.
Miller~\cite{miller_case_1996} attached time to each action and allowed overlapping actions with durations, along with a partial ordering of actions that allows describing two concurrent sequences of actions.
Pinto and Reiter~\cite{pinto_reasoning_1995} modeled durative actions with instantaneous \emph{start} and \emph{end} actions and continuous time to describe actions and events.
By distinguishing actual and possible situations, they define a total ordering on situations that allows expressing linear-time temporal properties within the \ac{SC}.
Finzi and Pirri~\cite{finzi_representing_2005} combine the \ac{SC} with temporal constraint reasoning, among others supporting \emph{processes} that have start and end actions and explicit start times.
Several approaches~\cite{miller_case_1996,reiter_natural_1996} allow modeling continuous processes, where a fluent continuously changes its value between two situations.
The \emph{hybrid \acl{SC}}~\cite{batusov_hybrid_2019} combines the \ac{SC} with hybrid systems by embedding \emph{hybrid automata}~\cite{alur_hybrid_1993}.

The \emph{event calculus}~\cite{kowalski_logic-based_1986,mueller_event_2008} uses an explicit representation of discrete or continuous time, where \emph{events} occur at certain points on the timeline.
In contrast to the \ac{SC}, formulas do not refer to situations, but to points in time, and it uses circumscription to solve the frame problem.
A comparison of the two formalisms can be found in \cite{van_belleghem_relation_1997}.
As there is no action programming language such as \golog in the event calculus, we focus on the \ac{SC}.

Timed automata (\acsp{TA})~\cite{alur_theory_1994} extend finite automata with continuous time.
While reachability is decidable for the original model, several extensions render it undecidable~\cite{alur_decision_2004}.
For \acp{TA} with \emph{diagonal} constraints of the form $x-y \bowtie c$ (in contrast to $x \bowtie c$), reachability remains decidable~\cite{alur_theory_1994},
while it is undecidable with \emph{additive constraints} ($x+y \bowtie c$)~\cite{berard_timed_2000}.
Allowing clock resets to non-zero values also affects decidability: with diagonal-free constraints, resets of clocks such as $x\coloneqq x+1$ preserve decidability, while diagonal constraints render reachability undecidable.
Updates of the form $x \coloneqq x -1$ always yield undecidability~\cite{bouyer_updatable_2004}.
As decidability for timed automata is well-studied, it is reasonable to adopt this model of time for the \ac{SC}.

\section{The Situation Calculus}\label{sec:foundations}
The \acf{SC}~\cite{mccarthy_situations_1963,reiter_knowledge_2001} is a sorted predicate logic for reasoning about dynamically changing worlds.
All changes in the world are caused by \emph{actions}, which are terms in the language.
A \emph{situation} is a possible history of the world, where the distinguished function symbol $\doop(\alpha, \sigma)$ describes the resulting situation of doing action $\alpha$ in situation $\sigma$ and
the constant $S_0$ denotes the initial situation where no actions have yet occurred.
Besides actions and situations, the sort \emph{objects} is used for terms describing objects of the world.
\emph{Fluents} are relational or functional symbols that describe possibly changing properties of the world and that have a situation as their last argument.
Relational and functional symbols that remain  unchanged between situations are called \emph{rigid} and do not have a situation argument.
We denote relational symbols with upper-case letters, e.g., $\rfluent{isFull}(p, s)$, and functional symbols with lower-case letters, e.g., $\ffluent{counter}(s)$.
Apart from the situation argument, all arguments of a fluent are of type object.
We assume that there is a finite set of fluent symbols \fluents and
a finite number of action types.
In contrast to the usual definition, we also assume a \emph{finite set of objects} \objects, for which we require the unique name assumption and domain closure.
For an action type $A$, we write $|A|$ for the number of arguments of $A$.
In the following, $s$ denotes a situation variable, $\sigma$ a ground situation term, $a$ an action variable, $\alpha$ a ground action term, $A$ an action type, $p$ and $o$ object variables, $\rho$ a ground object term, and $\vec{x}$ a tuple of variables $x_1, \ldots, x_k$ of the corresponding sort (similarly for terms).
We also write $\phi_1 \supset \phi_2$ for $\neg \phi_1 \vee \phi_2$ and $\phi_1 \equiv \phi_2$ for $\phi_1 \supset \phi_2 \wedge \phi_2 \supset \phi_1$, where $\equiv$ has lowest operator precedence and free variables are assumed to be universally quantified from the outside, e.g., $F(\vec{o}, s) \equiv \phi_1 \vee \phi_2$ stands for $\forall \vec{o}, s.(F(\vec{o}, s) \equiv (\phi_1 \vee \phi_2))$.
Finally, we write $\doop([\alpha_1; \ldots; \alpha_k], S_0)$ for $\doop(\alpha_k, \doop(\ldots, \doop(\alpha_1, S_0)))$.
A formula is \emph{uniform in $s$} if it does not mention any situation terms other than $s$ and does not mention $\poss$ (see \cite{reiter_knowledge_2001} for a formal definition).
A \emph{situation-suppressed formula} is a formula where the situation argument of each fluent is omitted.
For a formula $\phi$ and a situation term $\sigma$, we write $\phi[\sigma]$ for the formula obtained by replacing each occurrence of $s$ with $\sigma$.
Similarly, for a situation-suppressed formula $\phi$, $\phi[\sigma]$ is like $\phi$ with every fluent augmented by the situation argument $\sigma$.

The \ac{SC} has been extended to represent time by adding an additional real-valued sort \emph{time} with the standard interpretation of the real numbers and their operations (addition, subtraction, etc.) and relations ($<$, $\leq$, etc.) and augmenting each action term $A(\vec o, t)$ with an additional time argument $t$, which specifies the time point when the action occurs~\cite{pinto_reasoning_1995,reiter_knowledge_2001}.
An additional function symbol $\atime(\alpha)$ is used to describe the starting time of the action situation, i.e., $\atime(A(\vec o, t)) \eqdef t$.
This allows axiomatizing the starting time of a situation with $\sstart(\doop(a,s)) = \atime(a)$ and \emph{processes} with a duration that is defined by its start and end action \cite{finzi_representing_2005}.
With real numbers as an additional sort and independent of whether actions have an execution timepoint, we may also allow \emph{real-valued fluent functions}.
As we will see in \autoref{sec:reachability}, having either durative processes or real-valued fluent functions without any constraints renders the reachability problem undecidable.

\subsection{Basic Action Theories}

A \acf{BAT}~\cite{pirri_contributions_1999} is an axiomatization of a domain and describes all available actions with precondition axioms and successor state axioms. In addition to the domain-specific axioms, a \ac{BAT} also contains the foundational axioms of the \ac{SC}.
Similar to \cite{de_giacomo_bounded_2012}, we include domain closure axioms for action types and objects. Here, we axiomatize a finite set of objects.
\begin{definition}
    A \acf{BAT} is a set of sentences $\bat = \bat_0 \cup \bat_\text{poss} \cup \bat_\text{ssa} \cup \bat_{ca} \cup \bat_{co} \cup \Sigma$ where: 
    \begin{itemize}
        \item $\bat_0$ is a description of the initial situation with a set of sentences uniform in $S_0$.
        Here, we assume complete information, i.e., for every atomic (fluent or rigid) formula $\phi$ uniform in $S_0$, either $\bat \models \phi$ or $\bat \models \neg \phi$.
        \item $\bat_\text{poss}$ are precondition axioms of the form
         $
            \poss(A(\vec{o}), s) \equiv \phi_A(\vec{o}, s)
         $,
          one per action type $A \in \actions$ and where $\phi_A(\vec{o}, s)$ is a formula uniform in $s$ that describes when the action $A(\vec{o})$ can be performed.
        \item $\bat_\text{ssa}$ are \acfp{SSA} describing action effects. For each relational fluent  $F \in \fluents$, $\bat_{\text{ssa}}$ contains an axiom of the form
         $ 
            F(\vec{o}, \doop(a, s)) \equivspace \phi_F(\vec{o}, a, s)
         $, 
          where $\phi_F(\vec{o}, a, s)$ is a formula uniform in $s$ that describes how the fluent $F(\vec{o})$ changes if some action $a$ is performed.
          For each functional fluent $f \in \fluents$, $\bat_{\text{ssa}}$ contains an axiom of the form
          $ 
            f(\vec{o}, \doop(a, s)) = y \equivspace \phi_f(\vec{o}, y, a, s)
          $, 
          where $\phi_f(\vec{o}, y, a, s)$ is a formula uniform in $s$ describing how the fluent $f(\vec{o})$ changes if an action $a$ is performed.
        \item $\bat_{ca}$ are domain closure and \acp{UNA}
        on actions \actions, e.g.,
        \begin{sizeddisplay}
        \begin{align*}
            &\forall a (\exists \vec{o}_1(a = A_1(\vec{o}_1)) \vee \ldots \vee \exists \vec{o}_n(a = A_n(\vec{o}_n)))
            \\
            &\forall \vec{o}_i \vec{o}_j(A_i(\vec{o}_i) \neq A_j(\vec{o}_j)) \text{ for all } i \neq j
            \\
            &\forall \vec{o}_1 \vec{o}_1'(A_1(\vec{o}_1) = A_1(\vec{o}_1') \supset \vec{o}_1 = \vec{o}_1') \wedge \ldots
             \wedge \forall \vec{o}_{n} \vec{o}_n'(A_n(\vec{o}_n) = A_n(\vec{o}_n') \supset \vec{o}_n = \vec{o}_n')
        \end{align*}
        \end{sizeddisplay}
        \item $\bat_{co}$ are domain closure and \acp{UNA} on object constants \objects, e.g.,
        $
        \forall o (o = \rho_1 \vee o = \rho_2 \vee \ldots) \wedge \rho_1 \neq \rho_2 \wedge \ldots
        $.
        \item $\Sigma$: foundational, domain-independent axioms, possibly including $\atime$ and $\sstart$ from above and including the relation $\sqsubset$ that provides an ordering relation on situations.
    \end{itemize}
\end{definition}

We call \bat a \emph{timed \ac{BAT}} if it includes axioms for $\atime$, $\sstart$, etc.\ from above and we say \bat is \emph{real-valued} if it includes a functional fluent of sort \emph{time}.
Note that these are two independent properties: We may have occurrence times for actions without having any real-valued fluents and we may use real-valued fluents without assigning any time values to actions.
In the following, we assume that situation terms only mention action types from \actions and object terms from \objects.
Following~\cite{reiter_knowledge_2001}, a situation $s$ is \emph{executable}, written $\exec(s)$, if every action performed to reach $s$ is possible in the situation where it occurred:
$
  \exec(s) \eqdef \doop(a, s') \sqsubseteq s \supset
  \poss(a, s')
$.

\section{Reachability}\label{sec:reachability}

Generally speaking, reachability asks whether there is a reachable situation where some property holds.
Formally, given a \ac{BAT} \bat and a formula $\phi$, $\phi$ is \emph{reachable} if there is an executable situation where $\phi$ holds, i.e., if $\bat \models \exists s \exec(s) \wedge \phi[s]$.
Many problems such as planning or verification of safety properties can be formulated as a reachability problem.
It is therefore desirable to have a sound and complete method to decide reachability. 
However, depending on the expressivity of the underlying framework, the reachability problem may become undecidable.
It is easy to see that for a given \ac{BAT} over a finite set of objects that does not make use of time, the reachability problem is decidable.
On the other hand, if we include fluents of sort \emph{time} and allow arbitrary arithmetic operations on those fluents, then the problem is no longer decidable:\footnote{Full proofs can be found in Appendix~\ref{sec:proofs}.}
\begin{theoremE}\label{thm:real-valued-undecidability}
  The reachability problem for a real-valued \ac{BAT} is undecidable.
\end{theoremE}
\begin{proofidea}
  By reduction of the halting problem for \acp{2CM}, which is undecidable \cite{minsky_computation_1967}.
  Given \iac{2CM} $\mathcal{M}$, we can construct \iac{BAT} $\bat_{\mathcal{M}}$ such that $\bat_{\mathcal{M}} \models \exists s.\, \exec(s) \wedge \rfluent{halt}(s)$ iff $\mathcal{M}$ halts.
\end{proofidea}
\begin{proofE}
Let $\mi{Incrs} = \{ (p, c, q)_i \}_i$ be the finite set of increment instructions, where $p$ is the instruction label, $c \in \{ 1, 2 \}$ is the index of the counter to be incremented, and $q$ is the next instruction label.
Similarly, let $\mi{Decrs} = \{ (p, c, q, r)_i \}_i$ be the finite set of decrement instructions, where $p$ is the instruction label, $c \in \{ 1, 2 \}$ is the index of the counter to be decremented, $q$ is the jump instruction if the condition is true, and $r$ is the jump instruction otherwise.
We define a \ac{BAT} $\bat_\mathcal{M}$ corresponding to a \ac{2CM} $\mathcal{M}$ as follows:
We use the unary relational fluent $\rfluent{next}(i)$ to describe the next instruction $i$, two nullary functional fluents $f_1, f_2$ for counter values (we omit parentheses and write $f_1$ instead of $f_1()$), and the nullary relational fluent \rfluent{halt} which is true iff $\mathcal{M}$ halted.
We define the \ac{BAT} $\bat_{\mathcal{M}}$ as follows.
Initially, both counting fluents are set to zero and the next instruction is $s_0$: $\bat_0 = \{ f_1 = 0, f_2 = 0, \rfluent{next}(i) \equiv i = s_0 \}$.
An action is possible if it is the next instruction: $\poss(i) \equiv \rfluent{next}(i)$.
The \ac{SSA} for each counting fluent $f_i \in \{ f_1, f_2 \}$ works as follows:
If the last instruction was an increment instruction from $\mi{Incrs}$ and it incremented $c_i$, then the fluent $f_i$ is incremented as well. Similarly, if the last instruction was a decrement instruction and the counter value was larger than zero, then the fluent is decremented, otherwise it remains unchanged:
    $f_i(\doop(a, s)) = n \equiv \bigvee_{(p, c, q) \in \mi{Incrs}} a = p \wedge c = i \wedge n = f_i + 1
      \vee \bigvee_{(p, c, q, r) \in \mi{Decrs}} a = p \wedge c = i
      \wedge (f_i > 0 \wedge n = f_i -1 \vee f_i = 0 \wedge n = 0)
      \vee f_i = n \wedge \bigwedge_{(p, c, q) \in \mi{Incrs}} (a \neq p \vee c \neq i)
      \wedge \bigwedge_{(p, c, q, r)\in\mi{Decrs}} (a \neq p \vee c \neq i)$.
The \ac{SSA} for \rfluent{next} works as follows:
If the last instruction was an increment instruction that sets the next instruction to $q$, then set $\rfluent{next}$ accordingly. Similarly, if the last instruction was a decrement instruction, then branch on the value of the corresponding counter (i.e., whether $f_i > 0$) and set the next instruction accordingly:
    $\rfluent{next}(i) \equiv \bigvee_{(p, c, q) \in \mi{Incrs}} a = p \wedge i = q
        \vee \bigvee_{(p, c, q, r) \in \mi{Decrs}} a = p
        \wedge\big((c = 1 \wedge f_1 > 0 \vee c = 2 \wedge f_2 > 0) \wedge i = q
      \vee (c = 1 \wedge f_1 = 0 \vee c = 2 \wedge f_2 = 0) \wedge i = r\big)$.

    By construction, $\bat_\mathcal{M} \models \exists s \exec(s) \wedge \rfluent{halt}(s)$ iff the \ac{2CM} reaches \textbf{HALT}.
\end{proofE}

Intuitively, the problem is that we may use time fluents as counters for \acp{2CM}. These counters may be unbounded because time fluents are not bounded.
A natural restriction is thus adding an upper bound $b$ to all fluents of sort \emph{time}: Given a bound $b \in \mathbb{Q}$ and \iac{BAT} \bat with $\fluents_\text{time}$ denoting all functional fluents of sort time, we express that all fluents are bounded by $b$ with
$\textrm{Bounded}_b(s) \eqdef \forall s'.\, s' \sqsubseteq s \supset \bigwedge_{f \in \fluents_\text{time}} \forall \vec{o}.\, f(\vec{o}, s') \leq b$.
We can now define the \emph{time-bounded reachability problem}:
Given a property $\phi$ and a bound $b \in \mathbb{Q}$, does $\bat \models \exists s \exec(s) \wedge \textrm{Bounded}_b(s) \wedge \phi[s]$ hold?
Unfortunately, this restriction is not sufficient to accomplish decidability:
\begin{theoremE}\label{thm:bounded-undecidability}
    The time-bounded reachability problem for a real-valued \ac{BAT} and a time bound $b \geq 2$ is undecidable.
\end{theoremE}
\begin{proofidea}
Again by reduction of the halting problem for \acp{2CM}. A counter value $u$ is encoded by a fluent value $2^{1-u}$. A counter is decremented by doubling and incremented by halving the corresponding fluent and tested for zero by comparing the corresponding fluent to $2$.
\end{proofidea}
\begin{proofE}
The construction is the same as in \autoref{thm:real-valued-undecidability}, except that counting fluents work differently:
A counter value $u$ is encoded by a fluent value $2^{1-u}$.
Hence, every increment of $f_i$ is replaced by an update to $\frac{f_i}{2}$ and every decrement of $f_i$ is replaced by an update to $2f_i$.
Furthermore, a comparison $f_i = 0$ is replaced by $f_i = 2$ and $f_i > 0$ is replaced by $f_i < 2$.
Hence, the initial situation is
$\bat_0 = \{ f_1 = 2, f_2 = 2, \rfluent{next}(i) \equiv i = s_0 \}$.
The precondition axiom is: $\poss(i) \equiv \rfluent{next}(i)$.
The \ac{SSA} for each counting fluent $f_i \in \{ f_1, f_2 \}$ is defined as:
$f_i(\doop(a, s)) = n \equiv \bigvee_{(p, c, q) \in \mi{Incrs}} a = p \wedge c = i \wedge n = \frac{f_i}{2}
\vee \bigvee_{(p, c, q, r) \in \mi{Decrs}} a = p \wedge c = i
\wedge (f_i < 2 \wedge n = 2f_i \vee f_i = 2 \wedge n = 2)
\vee f_i = n \wedge \bigwedge_{(p, c, q) \in \mi{Incrs}} (a \neq p \vee c \neq i)
\wedge \bigwedge_{(p, c, q, r) \in \mi{Decrs}} (a \neq p \vee c \neq i)$.
The SSA for \rfluent{next} is defined as:
$\rfluent{next}(i) \equiv \bigvee_{(p, c, q) \in \mi{Incrs}} a = p \wedge i = q
  \vee \bigvee_{(p, c, q, r) \in \mi{Decrs}} a = p
  \wedge \big((c = 1 \wedge f_1 < 2 \vee c = 2 \wedge f_2 < 2) \wedge i = q
  \vee (c = 1 \wedge f_1 = 2 \vee c = 2 \wedge f_2 = 2) \wedge i = r\big)
$

By construction, $\bat \models \exists s \exec(s) \wedge \textrm{Bounded}_2(s) \wedge \rfluent{halt}(s)$ iff the \ac{2CM} reaches the instruction \textbf{HALT}.
\qedhere
\end{proofE}

These results suggest that we need to restrict the arithmetic operators on time fluents such that time may not change its value arbitrarily, but is monotonically increasing.
However, this is not sufficient;
even if we do not allow real-valued fluents but instead only consider actions with time arguments and processes with start times \cite{finzi_representing_2005}, the reachability problem remains undecidable:
\begin{theoremE}\label{thm:undecidable-bat-no-real-fluents}
    The reachability problem for timed \acp{BAT} without real-valued functional fluents is undecidable.
\end{theoremE}
\begin{proofidea}
  By reduction of the language emptiness problem for \acp{TA} with 4 clocks $c_1, \ldots, c_4$ extended with additive clock constraints of the form $c_1 + c_2 = 1$, which is undecidable~\cite{berard_timed_2000}.
  For a \ac{TA} \ta, we can construct \iac{BAT} and a formula $\phi$ such that $\bat \models \exists s.\, \exec(s) \wedge \phi[s]$ iff the language of \ta is non-empty.
  The key idea is to model each clock as a separate process, where a clock reset starts the process, defining the clock value of every clock in every situation.
\end{proofidea}
\begin{proofE}
  By reduction of the language emptiness problem for \acp{TA} with 4 clocks extended with additive clock constraints of the form $c_1 + c_2 = 1$, which is undecidable \cite{berard_timed_2000}.
  Let $\mathcal{A} = (L, l_0, L_F, \Sigma_\mathcal{A}, X, E)$ be such \iac{TA}.
  For an edge $e = (l, a, g, Y, l')$, we write $l_e, g_e, Y_e, l'_e$ for its components, and for each clock $x \in X$, we write $R_x \eqdef \{ e \in E \mid x \in Y_e \}$ for the set of edges resetting $x$.
  We define a \ac{BAT} $\bat_\mathcal{A}$ corresponding to $\mathcal{A}$ as follows.
  For each edge $e \in E$, there is one action type $E_e(t)$ with occurrence time $t$.
  For each location $l \in L$, there is a relational fluent $\mi{Loc}_l(s)$.
  Finally, for each clock $x \in X$, there is a \emph{process} $\mi{Reset}_x(t, s)$, where $t$ is an argument of sort time.
  Let $\sstart(S_0) = 0$.
  Initially, $\mathcal{A}$ is in location $l_0$, and every clock s was last reset at time 0: $\mi{Loc}_{l_0}(S_0)$, $\neg \mi{Loc}_l(S_0)$ for each $l \neq l_0$, $\mi{Reset}_x(t, S_0) \equiv t = 0$ for each $x \in X$.
  For a clock constraint $g$ and variable $t$ of sort time, the formula $\mu_g(t, s)$ is defined by
  \begin{itemize}
    \item $\mu_{x \bowtie r}(t, s) \eqdef \exists t_x. \mi{Reset}_x(t_x, s) \wedge t - t_x \bowtie r$
    \item $\mu_{x+y = 1}(t, s) \eqdef \exists t_x, t_y. \mi{Reset}_x(t_x, s) \wedge \mi{Reset}_y(t_y, s) \wedge (t - t_x) + (t - t_y) = 1$
    \item $\mu_{g_1 \wedge g_2} \eqdef \mu_{g_1}(t, s) \wedge \mu_{g_2}(t, s)$
  \end{itemize}
  An edge action is possible iff time does not decrease, $\mathcal{A}$ is in the source location, and the guard holds at the occurrence time:
  $\poss(E_e(t), s) \equiv t \geq \sstart(s) \wedge \mi{Loc}_{l_e}(s) \wedge \mu_{g_e}(t, s)$.
  The SSA for $\mi{Loc}_l$ states that $\mathcal{A}$ is in location $l$ after an action iff that action is an edge with target $l$: $\mi{Loc}_l(\doop(a, s)) \equiv \bigvee_{e \in E, l'_e = l} \exists t. a = E_e(t)$.
  The SSA for $\mi{Reset}_x$ states that the last reset time of $x$ is the occurrence time of the current action if that action resets $x$, and is unchanged otherwise:
  $\mi{Reset}_x(t, \doop(a, s)) \equiv \bigvee_{e \in R_x} a = E_e(t) \vee \mi{Reset}_x(t, s) \wedge \bigwedge_{e \in R_x} \forall t'. a \neq E_e(t')$.
  By construction, $\bat_\mathcal{A} \models \exists s. \exec(s) \wedge \bigvee_{l \in L_F} \mi{Loc}_l(s)$ iff there is an accepting timed word of $\mathcal{A}$.
\end{proofE}
Together, Theorems \ref{thm:real-valued-undecidability} to \ref{thm:undecidable-bat-no-real-fluents} rule out the natural ways of modeling dense time in the \ac{SC} while retaining decidability.
Reachability is undecidable if fluents range over the reals and may be manipulated arithmetically (\autoref{thm:real-valued-undecidability}); bounding the range of these fluents does not help (\autoref{thm:bounded-undecidability}); and removing real-valued functional fluents altogether, such that time only occurs through timed actions, does not retain decidability either (\autoref{thm:undecidable-bat-no-real-fluents}).
In each case, the problem is not continuous time itself, but allowing time values to be combined arithmetically.
This motivates the model of \autoref{sec:clocks-in-sitcalc}, which keeps dense time but restricts the admissible operations by only allowing resets, comparisons against natural numbers, and uniform progression of time.
\section{Clocks in the Situation Calculus}\label{sec:clocks-in-sitcalc}

In this section, we propose an alternative approach for modeling time in the \ac{SC} that maintains decidability.
We adopt methods from timed automata theory by defining \emph{clocked \acp{BAT}} and show that the reachability problem for clocked \acp{BAT} is decidable.
In particular, we introduce \emph{clocks} as real-valued functional fluents with restricted \acp{SSA} that guarantee uniform passing of time and that allow an action to reset a clock to zero.
Furthermore, we require that the \ac{BAT} uses a distinguished action \action{wait} that causes time to pass.
Apart from \action{wait}, all actions are instantaneous, i.e., they do not have a time argument and there is no delay between the previous and the current action.
This allows using clocks both in action preconditions and in successor state axioms.
We restrict formulas mentioning clocks to comparisons to natural numbers:
\begin{definition}
We distinguish the following types of formulas:
\begin{enumerate}
    \item A formula $\phi$ is a \emph{clock comparison} if it is of the form $f(\vec{o}, s) \bowtie v$ or $v \bowtie v'$, where $f \in \fluents$ is some functional fluent, $v, v' \in \naturals$, and $\operatorname{\bowtie} \in \{ <, \leq, =, \geq, > \}$.
    \item A formula $\phi$ is called \emph{clocked} if every atomic sub-formula mentioning a term of sort \emph{time} is a clock comparison.
    \item A formula $\phi$ is called \emph{time-independent} if it does not mention any term of sort \emph{time}.
\end{enumerate}
\end{definition}
We call a clock comparison \emph{$\maxconst$-bounded} if it only compares against constants $v, v' \leq \maxconst$.
A clocked formula is $\maxconst$-bounded if each occurring clock comparison is $\maxconst$-bounded.
As we will see, these restrictions allow us to apply \emph{regionalization}, which defines an equivalence relation on clock values.
Based on regionalization, we will define a \emph{time-abstract bisimulation}, resulting in a \emph{finite} abstraction of the otherwise infinite tree of situations, which permits a decidable procedure for the reachability problem.
We start by introducing clocks to \acp{BAT}:
\begin{definition}
\label{def:clocked-bat}
    A \ac{BAT} \bat is called a \emph{clocked BAT} if:
    \begin{enumerate}
        \item Every functional fluent $f$ is real-valued, i.e., of sort \emph{time};
        \item For each functional fluent $f$, $\bat_0$ contains an axiom $f(\vec{o}, S_0) = 0$;
        \item For each functional fluent $f$, $\bat_{\text{ssa}}$ contains an axiom of the form
        \begin{sizeddisplay}
          \begin{multline*}
            f(\vec{o}, \doop(a, s)) = y \equivspace
            \exists t.\, a = \action{wait}(t) \wedge y = f(\vec{o}, s) + t
            \\ \vee \neg \exists t.\, a = \action{wait}(t) \wedge (\phi_f(\vec{o}, a, s) \wedge y = 0
            \vee \neg \phi_f(\vec{o}, a, s) \wedge y = f(\vec{o}, s))
          \end{multline*}
        \end{sizeddisplay}
          where $\phi_f(\vec{o}, a, s)$ is a time-independent formula uniform in $s$
          describing if the action $a$ resets the value of $f(\vec{o})$;
        \item For every relational fluent $R$, $\bat_{\text{ssa}}$ contains an axiom of the form
        $R(\vec{o}, \doop(a, s)) \equiv \phi_R(\vec{o}, a, s)$ where $\phi_R(\vec{o}, a, s)$ is a clocked formula uniform in $s$.
        \item Every precondition axiom in $\bat_{\text{poss}}$ is of the form
            $
                \poss(A(\vec{o}), s) \equiv \phi_A(\vec{o}, s)
            $
        such that $\phi_A(\vec{o}, s)$ is a clocked formula uniform in $s$;
        \item Waiting for some non-negative amount of time is always possible: $\poss(\wait(t), s) \equiv t \geq 0$;
    \end{enumerate}
\end{definition}
We call a clocked BAT \bat \emph{$\maxconst$-bounded} if each clocked formula occurring in \bat is $\maxconst$-bounded.
Note that \bat is always $\maxconst$-bounded for some $\maxconst \in \naturals$, because \bat consists of finitely many axioms.
For the sake of simplicity, we assume that all functional fluents are of domain \emph{time} and are initialized to zero.
While this is not strictly necessary, it simplifies the treatment of functional fluents later on.
The restricted form of the \acp{SSA} of functional fluents enforces a behavior similar to clocks in timed automata:
An action $\action{wait}(t)$ increments each clock by the value $t \in \realpos$; all other actions either leave the value unchanged or reset it to zero.%
\footnote{
This separation allows us to use clock comparisons in precondition axioms and \acp{SSA}. If we allowed time to pass with every action, then the precondition axiom would refer to the time of the \emph{previous} situation, as time is only incremented after the action has occurred.
}
Hence, we will also call  these fluents \emph{clocks}.
We write \clocks for the set of all ground situation-suppressed clock terms, we denote clock variables with $c, c_1, \ldots$ and (possibly situation-suppressed) ground clock terms with $\omega, \omega_1, \ldots$.
Finally, we also restrict the usage of clocks in all axioms of a clocked \ac{BAT}: Both precondition and successor state axioms may only use clocked formulas, which restrict any operations on clocks to clock comparisons.
As we will see later on, this is an essential restriction for allowing regionalization.
Nevertheless, it is easy to show that clocked \acp{BAT} are expressive enough to represent \acp{TA}.
We use a variant of the coffee delivery robot~\cite{reiter_knowledge_2001} as a running example:
\begin{example}\label{ex:bat-coffee}
  We model a coffee delivery service that has the durative actions \action{brew} and \action{pour} and objects $\mi{Pot}$, $\mi{Mug}_1$, and $\mi{Mug}_2$.
  The robot needs to brew coffee and then pour it into the mugs.
  The fluents \brewing  and \pouring indicate that the robot is currently brewing (respectively pouring) coffee.
  After brewing, the pot \hasCoffee and after pouring into a mug, the mug \hasCoffee.
  The clock $c_{\action{brew}}$ tracks the duration of \action{brew}, while $c_{\mi{glob}}$ tracks global time.
  Brewing takes at least 1 time unit.
  If the robot brews coffee for at least 2 time units, the coffee is \rfluent{strong}.
  \begin{itemize}
      \item Initially, the robot is doing nothing, nothing is filled, the coffee is not strong, and all clocks are initialized to zero.
      Also, only $\mi{Mug}_1$ should be filled with strong coffee:
      \begin{sizeddisplay}
      \begin{gather*}%
            \neg\brewing(p, S_0), \neg\pouring(p, m, S_0),
            \neg \hasCoffee(o, S_0),
            \neg \isStrong(o, S_0),\\
            c_{\mi{brew}}(p, S_0) = 0, c_{\mi{glob}}(S_0) = 0,
            \rfluent{wantStrong}(o) \equiv o = \mi{Mug}_1
      \end{gather*}%
      \end{sizeddisplay}
      \item The robot may start a durative action if it is not doing anything and end it if it has started it before.
        Brewing takes at least 1 time unit.
        In order to pour the coffee, there must be coffee in the pot:
      \begin{sizeddisplay}
      \begin{align*}
          \poss(\sac{brew}(p),s) &\equivspace 
          \neg\brewing(p,s) \land \neg \exists m.\, \pouring(p,m,s)
          \\
          \poss(\eac{brew}(p),s) &\equivspace 
          \brewing(p,s)\wedge c_{\text{brew}}(p, s) \geq  1
          \\
          \poss(\sac{pour}(p,m),s) &\equivspace \hasCoffee(p,s) \wedge \neg \exists p', m'.\, \pouring(p',m',s) \land \neg \brewing(p,s)
          \\
          \poss(\eac{pour}(p,m),s) &\equivspace \pouring(p,m,s)
          \\
          \poss(\action{wait}(t), s) &\equivspace t \geq 0
      \end{align*}
      \end{sizeddisplay}
      \item
        The \acp{SSA} state that \action{brew} fills a pot with coffee and makes it strong if it took at least 2 time units:
      \begin{sizeddisplay}
      \begin{align*}
      \hasCoffee(o, \doop(a,s)) &\equivspace a = \eac{brew}(o) \lor \exists p.\, a = \eac{pour}(p, o) \lor \hasCoffee(o,s)
      \\
       \isStrong(o, \doop(a,s)) &\equivspace a = \eac{brew}(o) \land c_{\mi{brew}}(o,s) \geq 2 \\ 
       & \lor \exists p.\, a = \eac{pour}(p, o) \wedge \isStrong(p, s)
       \lor \isStrong(o,s)\\
      \brewing(p, \doop(a,s)) &\equivspace a = \sac{brew}(p) \lor \brewing(p, s) \land a \neq \eac{brew}(p)
      \\
      \pouring(p, m, \doop(a,s)) &\equivspace a = \sac{pour}(p,m) \lor \pouring(p, m, s) \land a \neq \eac{pour}(p,m)
      \\
      c_{\mi{brew}}(p, \doop(a, s)) = y &\equivspace
      \exists  t (a = \action{wait}(t) \wedge y = c_{\mi{brew}}(p, s) + t) \vee \Big(\neg \exists t. a = \action{wait}(t)
      \\
      & \wedge\big(a = \sac{brew}(p) \wedge y = 0 \vee a \neq \sac{brew}(p) \wedge y = c_{\mi{brew}}(p, s) \big)\Big)
        \\
        c_{\mi{glob}}(\doop(a, s)) = y &\equivspace \exists t.\, a = \action{wait}(t) \wedge y = c_{\mi{glob}}(s) + t
        \\
        & \vee \neg \exists t.  a = \action{wait}(t) \wedge y = c_{\mi{glob}}(s)
      \end{align*}
      \end{sizeddisplay}
      \item
        There are no other objects or actions:
        \begin{sizeddisplay}
        \begin{align*}
            &\forall o (o = \mi{Pot} \vee o = \mi{Mug}_1 \vee \ldots) \wedge \mi{Mug}_1 \neq \mi{Pot} \wedge \ldots
            \\
            &\forall a (\exists o.a = \sac{brew}(o) \vee \ldots
            \lor \exists t.\,a = \action{wait}(t)) 
            \\
            &\land \forall o_1,o_2.\,\sac{brew}(o_1) \neq \eac{brew}(o_2) \land \ldots
            \\
            &\wedge \forall o_1,o_2.\, \sac{brew}(o_1) = \sac{brew}(o_2) \supset o_1 = o_2 \wedge \ldots
        \end{align*}
        \end{sizeddisplay}
  \end{itemize}
\end{example}

\subsection{Regression}\label{sec:clocked-regression}
A fundamental reasoning task in the \ac{SC} is \emph{projection}, i.e., determining whether a sentence $\phi$ is true in a future situation.
A common method to solve projection is \emph{regression}~\cite{pirri_contributions_1999}, which reduces the formula $\phi$
to an equivalent formula $\reg{\phi}$ which only refers to the initial situation.
Given a \ac{BAT}, regression substitutes each fluent mentioned in $\phi$ by the right-hand side of the corresponding \ac{SSA} until every fluent only mentions $S_0$.
Regression is sound and complete and thus may be used for reducing queries about a future situation to the initial situation.
Here, we use a simplified variant of regression, where we require that all action terms are ground:
\begin{definition}
    A formula $\phi$ is \emph{regressable} iff
    \begin{enumerate}
        \item $\phi$ is first order (i.e., the second-order axioms $\Sigma$ are not regressable).
        \item Every term of sort \emph{situation} mentioned in $\phi$ has the form $\doop([\alpha_1, \ldots, \alpha_n], S_0)$, where $\alpha_1, \ldots, \alpha_n$ are ground  terms of sort \emph{action}.
        \item Every atom mentioning a clock $f$ is of the form $f(\vec{o}, s) \bowtie \tau$, where $\vec{o}$ are terms of sort object and $\tau \in \reals$.\footnote{In particular, we disallow quantifying over terms of sort \emph{time}.}
        \item For every atom of the form $\poss(a, s)$ mentioned in $\phi$, $a$ has the form $A(\vec{o})$ for some action type $A$.
        \item $\phi$ does not quantify over situations.
        \item $\phi$ does not mention the symbol $\sqsubset$, nor does it mention  any equality atom $s  = s'$ for situation terms $s, s'$.
    \end{enumerate}
\end{definition}
We extend the standard regression operator \regop with a regression rule for clock comparisons:
\begin{definition}
    Let $\phi = f(\vec{\rho}, \doop(\alpha, \sigma)) \bowtie r$ where $f \in \fluents$ is a clock and $r \in \reals$.
    By \autoref{def:clocked-bat}, there is \iac{SSA} of the following form:
    \begin{sizeddisplay}
        \begin{align*}
            f(&\vec{o}, \doop(a, s)) = y \equivspace
            \exists t.\, a = \action{wait}(t) \wedge y = f(\vec{o}, s) + t
            \\ &\vee \neg \exists t. a = \action{wait}(t) \wedge (\phi_f(\vec{o}, a, s) \wedge y = 0
            \vee \neg \phi_f(\vec{o}, a, s) \wedge y = f(\vec{o}, s))
    \end{align*}
    \end{sizeddisplay}
    Then $\reg{\phi}$ is defined as follows:
    \begin{enumerate}
        \item If $\alpha = \action{wait}(t)$ for some $t \in \realpos$, then $\reg{\phi} \eqdef \reg{f(\vec{\rho}, \sigma) \bowtie r'}$ with $r' = r - t$.
        \item Otherwise:
        $\reg{\phi} \eqdef \reg{\phi_f(\vec{\rho}, \alpha, \sigma) \wedge 0 \bowtie r
            \vee \neg \phi_f(\vec{\rho}, \alpha, \sigma) \wedge f(\vec{\rho}, \sigma) \bowtie r}$
    \end{enumerate}    
\end{definition}
We obtain a variant of the regression theorem:
\begin{theoremE}\label{thm:regression}
    Let $\phi$ be a regressable formula. Then $\reg{\phi}$ is uniform in $S_0$ and $\bat \models \phi$ iff $\bat_0 \cup \bat_c \models \reg{\phi}$.
\end{theoremE}
\begin{proofE}
    By induction on the maximal length of all situation terms in $\phi$ and structural sub-induction on $\phi$.
    \begin{description}
    \item[Base case.]
      Let $\sigma = S_0$. Then $\reg{\phi} = \phi$ and $\bat \models \phi$ iff $\bat_0 \cup \bat_c \models \phi$ as $\phi$ is uniform in $S_0$.
    \item[Induction step.]
      Let $\sigma = \doop(\alpha, \sigma')$.
      Here, we only consider the case of formulas of the form $\phi = f(\vec{\rho}, \sigma) \bowtie r$ and refer to \cite{pirri_contributions_1999} for the other cases.
      If $\alpha = \action{wait}(t)$ for some $t \in \realpos$,  then $\reg{f(\vec{\rho}, \sigma) \bowtie r} = \reg{f(\vec{\rho},  \sigma') \bowtie r'}$ with $r' = r - t$.
      By induction, $\bat_0 \cup \bat_c \models \reg{\phi_f(\vec{\rho}, \alpha, \sigma) \bowtie r'}$ iff $\bat \models \phi_f(\vec{\rho}, \alpha, \sigma) \bowtie r'$.
      Otherwise (if $\alpha$ is not \action{wait}), by the definition of the \ac{SSA} for $f$, $\bat \models f(\vec{\rho}, \doop(\alpha, \sigma)) = y \equiv \phi_f(\vec{\rho}, \alpha, \sigma) \wedge y = 0 \vee \neg \phi_f(\vec{\rho}, \alpha, \sigma) \wedge y = f(\vec{\rho}, \sigma)$
      and so $\bat \models f(\vec{\rho}, \doop(\alpha, \sigma)) \bowtie r \equiv \phi_f(\vec{\rho}, \alpha, \sigma) \wedge 0 \bowtie r \vee \neg \phi_f(\vec{\rho}, \alpha, \sigma) \wedge f(\vec{\rho}, \sigma) \bowtie r$.
      By induction, this holds iff $\bat_0 \cup \bat_c \models \reg{\phi_f(\vec{\rho}, \alpha, \sigma) \wedge 0 \bowtie r \vee \neg \phi_f(\vec{\rho}, \alpha, \sigma) \wedge f(\vec{\rho}, \sigma) \bowtie r}$ and so $\bat \models f(\vec{\rho}, \doop(\alpha, \sigma)) \bowtie r$ iff $\bat_0 \cup \bat_c \models \reg{f(\vec{\rho}, \doop(\alpha, \sigma)) \bowtie r}$.
  \end{description}
\end{proofE}

With this variant of regression, deciding whether a \ac{BAT} entails a regressable clocked formula is decidable:
\begin{theoremE}[][normal]\label{thm:regression-decidable}
    Let $\phi$ be a regressable clocked formula and \bat a clocked \ac{BAT} (over finite objects and with complete information). Then $\bat \models \phi$ is decidable.
\end{theoremE}
\begin{proofE}
    As regression is sound and complete, it is sufficient to consider $\bat_0 \cup \bat_c \models \reg{\phi}$, which can be reduced to entailment in propositional logic:
    First, as \objects is finite, we can eliminate each quantifier in $\bat$ and $\phi$ by substitution, e.g., by substituting $\exists o. \phi'$ with $\bigvee_{o \in \objects} \phi'$.
    Second, note that regressing a clock comparison $f(\vec{\rho}, \sigma) \bowtie r$ results in a formula $f(\vec{\rho}, S_0) \bowtie r'$, where $r' \in \reals$.
    As all clock values are initialized to $0$, we can directly replace each such comparison by $\top$ or $\bot$, e.g., $f(\vec{\rho}, S_0) > 5$ is simplified to $\bot$.
    Finally, with the unique name assumption, we can interpret predicate symbols as propositional variables, thereby obtaining a propositional set of sentences $\bat_0'$ and a propositional formula $\phi'$ such that $\bat_0' \models \phi'$ iff $\bat_0 \cup \bat_c \models \phi$.
\end{proofE}

\subsection{Regionalization}
To show decidability of the reachability problem for clocked \acp{BAT}, we use an approach based on abstraction for timed automata inspired by~\cite{alur_theory_1994,ouaknine_decidability_2007}.
We construct a finite abstraction of the otherwise infinite tree of situations of a clocked \ac{BAT} $\bat$ using a time-abstract bisimulation relation over situations:
\begin{definition}[Time-abstract Bisimulation]\label{def:bisimulation}
   An equivalence relation $R \subseteq \situations \times \situations$ is a \emph{$\maxconst$-relative time-abstract bisimulation} on \situations if $(\sigma_1, \sigma_2) \in R$ implies
   \begin{enumerate}
      \item For every $\maxconst$-bounded clocked formula $\phi$ uniform in $s$: $\bat \models \phi[\sigma_1]$ iff $\bat \models \phi[\sigma_2]$.
      \item For every action type $A\neq\action{wait}$ and ground arguments $\vec{\rho}$, $(\doop(A(\vec{\rho}), \sigma_1), \doop(A(\vec{\rho}), \sigma_2)) \in R$.
      \item For every $\tau_1\in\realpos$, there is a $\tau_2\in\realpos$ such that $(\doop(\action{wait}(\tau_1), \sigma_1), \doop(\action{wait}(\tau_2), \sigma_2)) \in R$.
   \end{enumerate}
\end{definition}
Intuitively, two situations are bisimilar if they represent the same state of the world (i.e., they satisfy the same \maxconst-bounded clocked formulas) and if applying the same action in both situations results again in bisimilar situations.

In the following, we construct such a bisimulation by applying regionalization to situations.
Generally, a \emph{clock valuation over clocks $C$} is a function $\clockval: C \rightarrow \realpos$.
For situations, we can define clock valuations as follows:
\begin{definition}[Clock Valuation]
    Let $\bat$ be a clocked \ac{BAT} with (situation-suppressed) clocks $\clocks$ and $\sigma$ a situation.
    The clock valuation $\clockval_\sigma^\bat$ of $\sigma$ is the clock valuation over \clocks with $\clockval_\sigma^\bat(\omega) = \tau$ iff $\bat \models \omega[\sigma] = \tau$ for a ground clock term $\omega$.
\end{definition}
We will omit the superscript \bat if \bat is clear from context.
It is sometimes convenient to write a clock valuation \clockval as a set of pairs $\{ (c, v) \mid c \in C, v = \clockval(c) \} \subseteq C \times \realpos$.
We also write $\clockval + \tau$ for $\{ (c, v + \tau) \mid c \in C, v = \clockval(c) \}$ where all clocks are incremented by $\tau$.
We can show that every situation has a unique clock valuation:
\begin{lemmaE}
    If \bat is a clocked \ac{BAT} and $\sigma$ a situation, then $\clockval_\sigma$ always exists and is unique.
\end{lemmaE}
\begin{proofE}
    By definition of clocked \acp{BAT}, each clock is initialized to $0$.
    Furthermore, once the initial value is fixed, the \acp{SSA} uniquely define each clock value in all successor situations.
\end{proofE}

We use \emph{regionalization} to obtain an abstraction of clock valuations.
We briefly outline the intuition behind this idea:
Given a constant $\maxconst \in \naturals$ that is the maximal constant of sort \emph{time} mentioned in \bat and in the reachability query $\phi$, we can collect all clock values with a value larger than $\maxconst$ in a single class $\top$.
Furthermore, as clocks are only compared to natural numbers, for a single clock it suffices to  observe only integer parts and whether fractional parts differ from zero to derive a finite set of equivalence classes for clock valuations $v\in[0,\maxconst]$. 
In the presence of multiple clocks, the order between the fractional parts needs to be taken into account as well~\cite{alur_timed_1999}.
We define the \emph{fractional part} $\fract(v)$ of $v \in [0,\maxconst]\cup\{\top\}$ as $v - \lfloor v \rfloor$  if  $v \in [0, \maxconst]$ and zero otherwise.
This leads to the definition of clock regions:
\begin{definition}[Clock Regions]\label{def:regionalization}
  Given a maximal constant $\maxconst$, let $V = [0, \maxconst] \cup \{ \top \}$.
  We define the \emph{region equivalence} as the equivalence relation $\clockequiv$ on $V$ such that $u \clockequiv v$ if
  \begin{itemize}
    \item $u = v = \top$, or
    \item $u, v \neq \top$, $\lceil u \rceil = \lceil v\rceil$, and $\lfloor u \rfloor = \lfloor v \rfloor$.
  \end{itemize}
  A \emph{region} is an equivalence class of $\operatorname{\clockequiv}$.
  We extend region equivalence to clock valuations.
  Let $\clockval, \clockval'$ be two clock valuations over $\clocks$, then $\clockval$ and $\clockval'$ are region-equivalent, written $\clockval \regequiv \clockval'$ iff
  (1) for every $c \in \clocks$, $\clockval(c) \clockequiv \clockval'(c)$, and
  (2) for every $c, c' \in \clocks$, $\fract(\clockval(c)) \leq \fract(\clockval(c'))$ iff $\fract(\clockval'(c)) \leq \fract(\clockval'(c'))$.
\end{definition}

Note that for a fixed \maxconst, there is only a finite number of regions.
Based on region equivalence wrt.\ \maxconst, we can now define an equivalence relation on situations:
\begin{definition}\label{def:situationEquivRelation}
    Let $\operatorname{\tabisim} \subseteq \situations \times \situations$ be an equivalence relation such that $\sigma_1 \tabisim \sigma_2$ iff
    \begin{enumerate}
        \item for every fluent $R \in \fluents$, $\bat \models R(\vec{\rho}, \sigma_1)$ iff $\bat \models R(\vec{\rho}, \sigma_2)$,
        \item there is a bijection $f: \clockval_{\sigma_1} \rightarrow \clockval_{\sigma_2}$ such that
        \begin{enumerate}
            \item if $f(c_1, v_1) = (c_2, v_2)$, then $c_1 = c_2$ and $v_1 \clockequiv  v_2$
            \item if $f(c_1, v_1) = (c_2, v_2)$ and $f(c_1', v_1') = (c_2', v_2')$, then $\fract(v_1) \leq \fract(v_1')$ iff $\fract(v_2) \leq \fract(v_2')$.
        \end{enumerate}
    \end{enumerate}
\end{definition}
Two situations are equivalent if they satisfy the same fluents and agree on the clock regions.
For a situation $\sigma$, we write $[\sigma]_{\tabisim}$ for the equivalence class of $\sigma$ wrt.\ \tabisim. We denote the set of equivalence classes resulting from \tabisim as $\situations/\tabisim$.
Equivalence directly extends to clocked formulas:
\begin{lemmaE}\label{lma:bisim:static-equivalence}
  Let $\sigma_1 \tabisim \sigma_2$ and let $\phi$ be a $\maxconst$-bounded clocked formula uniform in $s$. Then $\bat \models \phi[\sigma_1]$ iff $\bat \models \phi[\sigma_2]$.
\end{lemmaE}
\begin{proofE}
  Let $f: \clockval_{\sigma_1} \rightarrow \clockval_{\sigma_2}$ be a bijection witnessing $\sigma_1 \tabisim \sigma_2$.
  We show that $\bat \models \phi[\sigma_1]$ iff $\bat \models \phi[\sigma_2]$ by structural induction on $\phi$.
  \begin{itemize}
      \item Let $\phi = R(\vec{\rho}, s)$ for some fluent predicate $R$.
        By \autoref{def:bisimulation}, it directly follows that $\bat \models \phi[\sigma_1]$ iff $\bat \models \phi[\sigma_2]$.
      \item Let $\phi = \omega \bowtie \tau$ for some ground situation-suppressed clock term $\omega \in \clocks$ and $\tau \in \naturals$.
      Let $\clockval_{\sigma_1}(\omega) = v_1$ and $\clockval_{\sigma_2}(\omega) = v_2$.
      As $f$ is a witness for $\sigma_1 \tabisim \sigma_2$, it follows that $f(\omega, v_1) = (\omega, v_2)$ and $v_1 \clockequiv v_2$.
      By \autoref{def:regionalization}, $v_1 \bowtie \tau$ iff $v_2 \bowtie \tau$ and so $\bat \models \omega[\sigma_1] \bowtie \tau$ iff $\bat \models \omega[\sigma_2] \bowtie \tau$.
      \item The cases for Boolean connectives and quantification follow by induction.
      \qedhere
  \end{itemize}
\end{proofE}

We use a well-known result from \cite{alur_theory_1994} which states that it is always possible to maintain equivalence wrt. $\regequiv$ between clock valuations for a given time increment:
\begin{proposition}\label{prop:reg-successor}
  Let $\clockval_1$ and $\clockval_2$ be two clock valuations over a set of clocks \clocks such that $\clockval_1 \regequiv \clockval_2$.
  Then for all $\tau \in \realpos$ there exists a $\tau' \in \realpos$ such that $\clockval_1 + \tau \regequiv \clockval_2 + \tau'$.
\end{proposition}
Therefore, for two situations with $\sigma_1 \tabisim \sigma_2$ and any action $\action{wait}(\tau_1)$, it is always possible to find a $\tau_2$ such that $\doop(\action{wait}(\tau_1), \sigma_1) \tabisim \doop(\action{wait}(\tau_2), \sigma_2)$.
Based on this, we can show that \tabisim satisfies all criteria of \autoref{def:bisimulation}:
\begin{theoremE}\label{thm:equivRelation}
    $\tabisim$ is a time-abstract bisimulation.
\end{theoremE}
\begin{proofE}
  We show that \tabisim satisfies the criteria from \autoref{def:bisimulation}.
  Let $\sigma_1 \tabisim \sigma_2$ such that $f: \clockval_{\sigma_1} \rightarrow \clockval_{\sigma_2}$ is a bijection witnessing $\sigma_1 \tabisim \sigma_2$.
  \begin{enumerate}
      \item
      Let $\phi$ be a clocked formula uniform in $s$.
      It directly follows from \autoref{lma:bisim:static-equivalence} that $\bat \models \phi[\sigma_1]$ iff $\bat \models \phi[\sigma_2]$.
      \item Let $A\neq\action{wait}$ be some action type, $\vec{\rho}$ a ground tuple of arguments. Let $\sigma_1'=\doop(A(\vec{\rho}), \sigma_1)$ and $\sigma_2'=\doop(A(\vec{\rho}), \sigma_2)$. 
      We show that $\sigma_1' \tabisim \sigma_2'$:
      \begin{enumerate}
          \item Let $R \in \fluents$ be a relational fluent and $\pvec{\rho}'$ a ground tuple of arguments.
          We show that $\bat \models R(\pvec{\rho}', \sigma_1')$ iff $\bat \models R(\pvec{\rho}', \sigma_2')$.
          There is \iac{SSA} for $R$ of the form $R(\vec{o}, \doop(a, s)) \equiv \phi_R(\vec{o}, a, s)$, where $\phi_R(\vec{o}, a, s)$ is a clocked formula uniform in $s$.
          As $\sigma_1 \tabisim \sigma_2$, it directly follows from \autoref{lma:bisim:static-equivalence} that $\bat \models \phi_R(\pvec{\rho}', A(\vec{\rho}), \sigma_1)$ iff $\bat \models \phi_R(\pvec{\rho}', A(\vec{\rho}), \sigma_2)$ and therefore $\bat \models R(\pvec{\rho}', \sigma_1')$ iff $\bat \models R(\pvec{\rho}', \sigma_2')$.
          \item We construct a bijection $f': \clockval_{\sigma_1'} \rightarrow \clockval_{\sigma_2'}$ witnessing \tabisim.
          Let $c$ be a functional fluent and $\omega = c(\pvec{\rho}')$.
          There is \iac{SSA} of the form $
            c(\vec{o}, \doop(a, s)) = y \equivspace
            \exists t.\, a = \action{wait}(t) \wedge y = c(\vec{o}, s) + t
            \vee \neg \exists t.a = \action{wait}(t) \wedge (\phi_c(\vec{o}, a, s) \wedge y = 0 
            \vee \neg \phi_c(\vec{o}, a, s) \wedge y = c(\vec{o}, s))$.
          Thus, $\omega[\sigma_1'] = 0$ if $\bat \models \phi_c(\vec{\rho}, \alpha, \sigma_1)$ and $\omega[\sigma_1'] = \omega[\sigma_1]$ otherwise, similarly for $\omega[\sigma_2']$.
          Now, as $\phi_c(\vec{\rho}, a, s)$ is a clocked formula uniform in $s$, by \autoref{lma:bisim:static-equivalence}, $\bat \models \phi_c(\vec{\rho}, \alpha, \sigma_1)$ iff $\bat \models \phi_c(\vec{\rho}, \alpha, \sigma_2)$.
          Hence, for each $\omega$, we can set $f'(\omega, 0) = (\omega, 0)$ if $\bat \models \phi_c(\vec{\rho}, \alpha, \sigma_1)$ and $f'(\omega, \clockval_{\sigma_1}(\omega)) = (\omega, \clockval_{\sigma_2}(\omega))$ otherwise.
      \end{enumerate}
      \item Let $\tau_1 \in \realpos$ and  $\sigma_1' = \doop(\action{wait}(\tau_1), \sigma_1)$.
      As $\clockval_{\sigma_1} \regequiv \clockval_{\sigma_2}$, it follows with \autoref{prop:reg-successor} that there is some $\tau_2 \in \realpos$ such that $\clockval_{\sigma_1} + \tau_1 \regequiv \clockval_{\sigma_2} + \tau_2$.
      Let $\sigma_2' = \doop(\action{wait}(\tau_2), \sigma_2)$.
      We show that $\sigma_1' \tabisim \sigma_2'$:
      \begin{enumerate}
          \item Let $R \in \fluents$ be a relational fluent and $\pvec{\rho}'$ a ground tuple of arguments.
          We show that $\bat \models R(\pvec{\rho}', \sigma_1')$ iff $\bat \models R(\pvec{\rho}', \sigma_2')$.
          There is \iac{SSA} for $R$ of the form $R(\vec{o}, \doop(a, s)) \equiv \phi_R(\vec{o}, a, s)$, where $\phi_R(\vec{o}, a, s)$ is a clocked formula uniform in $s$.
          As $\sigma_1 \tabisim \sigma_2$, it directly follows from \autoref{lma:bisim:static-equivalence} that $\bat \models \phi_R(\pvec{\rho}', \action{wait}(\tau_1), \sigma_1)$ iff $\bat \models \phi_R(\pvec{\rho}', \action{wait}(\tau_2), \sigma_2)$ and therefore $\bat \models R(\pvec{\rho}', \sigma_1')$ iff $\bat \models R(\pvec{\rho}', \sigma_2')$.
          \item
          Let $c$ be a functional fluent and $\omega = c(\pvec{\rho}')$.
          By definition of the \ac{SSA} for  $c$, it follows that $\clockval_{\sigma_1'}(\omega) = \clockval_{\sigma_1}(\omega) + \tau_1$ and $\clockval_{\sigma_2'}(\omega) = \clockval_{\sigma_2}(\omega) + \tau_2$.
          We can set $f'(\omega, \clockval_{\sigma_1'}(\omega)) = (\omega, \clockval_{\sigma_2'}(\omega))$.
          As $\clockval_{\sigma_1} + \tau_1 \regequiv \clockval_{\sigma_2} + \tau_2$, $f'$ is a bijection witnessing \tabisim.
          \qedhere
      \end{enumerate}
  \end{enumerate}
\end{proofE}
Hence, it suffices to argue about the resulting equivalence classes to determine entailment with respect to a given $\bat$.
This enables us to construct a time-abstract transition system as a finite abstraction for the otherwise infinite tree of situations, where states collect $\tabisim$-equivalent situations connected by executable actions from $\bat$:
\begin{definition}[Time-abstract Transition System]
  Let \bat be a clocked \ac{BAT}.
  We define the TS $\absts = (Q, q_0, \rightarrow)$ as follows:
  \begin{itemize}
    \item The set of states $Q$ consists of equivalence classes of \tabisim, i.e., $Q \subseteq \situations / \tabisim$;
    \item The initial state is the equivalence class $q_0 = [S_0]_{\tabisim}$; 
    \item $[\sigma]_{\tabisim} \rightarrow [\sigma']_{\tabisim}$ if $\sigma'$ is an executable successor situation of $\sigma$, i.e., if $\sigma' = \doop(\alpha, \sigma)$ and $\bat \models \exec(\sigma')$.
  \end{itemize}
\end{definition}

It is easy to see that it is sufficient to consider \absts for checking reachable situations:
\begin{lemmaE}\label{lma:absts-reachability}
  $[\sigma]_{\tabisim}$ is reachable in \absts iff there is some $\sigma' \in [\sigma]_{\tabisim}$ with $\bat \models \exec(\sigma')$.
\end{lemmaE}
\begin{proofE}
    ~\\
    \textbf{$\Leftarrow$:}
    By induction on the number $n$ of \doop operators in $\sigma$.
    \begin{description}
      \item[Base case.]
        Let $n = 0$. Therefore, $\sigma = S_0$. By definition, $[S_0]_{\tabisim}$ is reachable in \absts.
      \item[Induction step.]
        Let $\sigma = \doop(\alpha, \sigma^*)$ such that $\bat \models \exec(\sigma)$.
        By induction, $[\sigma^*]_{\tabisim}$ is reachable from $[S_0]_{\tabisim}$ in \absts.
        By definition of \absts, $[\sigma^*]_{\tabisim} \rightarrow [\sigma]_{\tabisim}$ is a transition in \absts and so $[\sigma]_{\tabisim}$ is also reachable in \absts.
    \end{description}
    \textbf{$\Rightarrow$:}
    Follows immediately by the definition of \absts.
\end{proofE}

In contrast to the tree of situations, the time-abstract transition system \absts is finite:
\begin{lemmaE}\label{lma:absts-finite}
    The time-abstract transition system \absts has finitely many states.
\end{lemmaE}
\begin{proofE}
    As there are only finitely many fluents \fluents, objects \objects, and clock regions, \tabisim has finitely many equivalence classes.
    Also, as \objects and \actions are finite, each state only has finitely many successors.
    Therefore, the time-abstract transition system \absts has finitely many states.
\end{proofE}

Hence, the time-abstract transition system \absts is useful for deciding reachability: every reachable situation is also reachable in \absts, which is a finite transition system and can be explored exhaustively.
It remains to be shown that we can actually construct \absts.

\subsection{Decidability}

\begin{algorithm}[tb]
\centering
\caption{Computation of the time-abstract transition system \absts for a clocked \ac{BAT} $\bat$.}
\label{alg:regiontransitionsystem}
\smallfont
\begin{algorithmic}[0]
\State $\mi{Open} \gets \{ S_0 \}$;
$V \gets \{ [S_0]_{\tabisim} \}$;
$E \gets \emptyset$
\While{$\mi{Open} \neq \emptyset$}
    \State $\sigma \gets \Call{pop}{\mi{Open}}$
    \State $\mi{Acts} \gets \{ A(\vec{\rho}) \mid A \in \actions \setminus \{ \action{wait} \}, \vec{\rho} \in \mathcal{O}^{|A|} \}  \cup \{ \action{wait}(\tau) \mid \tau \in \Call{TSuccs}{\sigma, \maxconst} \}$\label{alg:absts:sym-acts}\label{alg:absts:wait-acts}
    \ForAll{$\alpha \in \mi{Acts}$}
        \If{$\bat_0 \cup \bat_c \models \reg{\poss(\alpha, \sigma)}$}
            \If{$[\doop(\alpha, \sigma)]_{\tabisim} \not\in V$}\label{alg:abs:check-v}
                \State $V \gets V \cup \{ [\doop(\alpha, \sigma)]_{\tabisim} \}$\label{alg:absts:add-v}
                \State $\mi{Open} \gets \mi{Open} \cup \{ \doop(\alpha, \sigma) \}$ 
            \EndIf
            \State $E \gets E \cup \{ ([\sigma]_{\tabisim}, [\doop(\alpha, \sigma)]_{\tabisim}) \}$\label{alg:absts:add-e}
        \EndIf
    \EndFor
\EndWhile
\State \Return $(V,[S_0]_{\tabisim},E)$
\end{algorithmic}
\end{algorithm}

\begin{algorithm}[thb]
\centering
\caption{Time successor computation for a situation $\sigma$.}
\label{alg:timesuccs}
\begin{algorithmic}[0]
\smallfont
\Function{TSuccs}{$\sigma, K$}
\State
$\clockval \gets \clockval_\sigma$;
$\mi{last} \gets 0$;
$\mi{Succs} \gets \{ \mi{last} \}$
\While{$\exists (c, v) \in \clockval:\: v \leq K$}
  \State $\mu \gets \max_{(c, v) \in \clockval} \fract(v)$
  \If{$\exists (c, v) \in \clockval:\: \fract(v) = 0$}
    $\mi{incr} \gets \frac{1-\mu}{2}$
  \Else
    $\;\mi{incr} \gets 1 - \mu$
  \EndIf
  \State $\mi{last} \gets \mi{last} + \mi{incr}$
  \State $\mi{Succs} \gets \mi{Succs} \cup \{ \mi{last} \}$
  \State $\clockval \gets \clockval + \mi{incr}$
\EndWhile
\State \Return $\mi{Succs}$
\EndFunction
\end{algorithmic}
\end{algorithm}

We now describe an algorithm for constructing \absts, which is shown in \autoref{alg:regiontransitionsystem}.
Starting with the initial state $[S_0]_{\tabisim}$, the algorithm iteratively adds a state $[\sigma]_{\tabisim}$ to \absts whenever $\sigma$ is reachable by a possible single action from an existing state, i.e., if $\sigma = \doop(\alpha, \sigma')$ and $\bat \models \poss(\alpha, \sigma')$ for some state $[\sigma']_{\tabisim}$ of \absts.
It does so in a breadth-first search manner by keeping a list of \emph{open} states $\mi{Open}$, which consists of states that need to be expanded.
The algorithm computes all possible successors by enumerating all possible ground action terms.
For the action \action{wait}, it includes a canonical timepoint $\tau$ as a representative for the equivalence class $[\clockval_\sigma + \tau]_{\regequiv}$.
These canonical time successors $\textsc{TSuccs}(\sigma,\maxconst)$ are computed using \autoref{alg:timesuccs} based on a well-known method \cite{ouaknine_decidability_2007}, which satisfies the following property:
\begin{proposition}\label{lma:tsucc}
  Let $\clockval$ be a clock valuation, $\tau \in \realpos$, and $\mi{Succs}$ the time successors computed by \autoref{alg:timesuccs}.
  Then there is a $\tau' \in \mi{Succs}$ such that $\clockval + \tau \regequiv \clockval + \tau'$.
\end{proposition}
The reachability problem is indeed decidable:
\begin{lemmaE}\label{lma:sitRelatableIsDecidable}
    Let $\sigma_1$ and $\sigma_2$ be two situations. Then it is decidable whether $\sigma_1 \tabisim \sigma_2$.
\end{lemmaE}
\begin{proofE}
    For every fluent $R \in \fluents$ and parameters $\vec{\rho}$,  $\bat \models R(\vec{\rho}, \sigma_1)$ iff $\bat_0 \cup \bat_c \models \reg{R(\vec{\rho}, \sigma_1)}$, similarly for $\sigma_2$.
    Following \autoref{thm:regression-decidable}, checking whether $\bat_0 \cup \bat_c \models \reg{R(\vec{\rho}, \sigma)}$ is decidable. 
    Furthermore, because \bat has finitely many objects and finitely many fluents, there are only finitely many fluent atoms that need to be checked.
    Finally, checking for the existence of a bijection $f: \clockval_{\sigma_1} \rightarrow \clockval_{\sigma_2}$ can be done by enumerating all possible bijections (of which there are finitely many) and checking whether all requirements are satisfied.
\end{proofE}

\begin{lemmaE}\label{lma:alg-returns-absts}
For a clocked \ac{BAT} \bat,
\autoref{alg:regiontransitionsystem} terminates and returns the time-abstract transition system \absts.
\end{lemmaE}
\begin{proofE}
    Let \absts  be the time-abstract TS of a \ac{BAT} \bat and $T$ the TS returned by \autoref{alg:regiontransitionsystem}.
    We show for every path $p$ starting in $[S_0]_{\tabisim}$ that $p$ is a path in \absts iff $p$ is a path in $T$.
    \\
    \textbf{$\Rightarrow$:}
        Let $p = [S_0]_{\tabisim} \rightarrow^* [\sigma]_{\tabisim}$ be a path in \absts.
        We show by induction on the length $n$ of $p$ that $p$ is a path in $T$.
        \begin{description}
          \item[Base case.]
            Let $n = 0$, i.e., $p$ is the empty path, which is also a path in $T$.
          \item[Induction step.]
            Let $p = [S_0]_{\tabisim} \rightarrow^* [\sigma^*]_{\tabisim} \rightarrow [\sigma]_{\tabisim}$.
            By induction, $[S_0]_{\tabisim} \rightarrow^* [\sigma^*]_{\tabisim}$ is a path in $T$.
            It remains to be shown that $[\sigma^*]_{\tabisim} \rightarrow [\sigma]_{\tabisim}$ is a transition in $T$.
            Clearly, $\sigma = \doop(\alpha, \sigma^*)$ for some $\alpha$.
            We distinguish two cases:
            \begin{enumerate}
                \item
                    Assume $\alpha = A(\vec{\rho})$ for some action type $A \in \actions \setminus \{ \action{wait} \}$.
                    $\alpha$ is added to $\mi{Acts}$ in line~\ref{alg:absts:sym-acts}.
                    By definition of \absts, $\bat \models \exec(\sigma)$ and so $\bat \models \poss(\alpha, \sigma^*)$.
                    Hence, $[\sigma]_{\tabisim} \in V$ (either it is added in line~\ref{alg:absts:add-v} or it was in $V$ before) and $([\sigma^*]_{\tabisim}, [\sigma]_{\tabisim})$ is added to $E$ in line~\ref{alg:absts:add-e}.
                \item
                    Otherwise, $\alpha = \action{\wait}(\tau)$ for some $\tau \in \realpos$.
                    By \autoref{lma:tsucc}, there is a $\tau' \in \mi{Succs}$ such that $\clockval_{\sigma^*} + \tau \regequiv \clockval_{\sigma^*} + \tau'$ and so $\alpha' = \action{wait}(\tau')$ is added to $\mi{Acts}$ in line~\ref{alg:absts:wait-acts}.
                    Also, $\doop(\alpha, \sigma^*) \tabisim \doop(\alpha', \sigma^*)$, and so $[\doop(\alpha, \sigma^*)]_{\tabisim} = [\doop(\alpha', \sigma^*)]_{\tabisim} = [\sigma]_{\tabisim}$.
                    Furthermore, as the precondition axiom of \action{wait} is $\poss(\action{wait}(t), s) \equiv t \geq 0$, it is clear that $\bat_0 \cup \bat_c \models \reg{\poss(\alpha', \sigma)}$.
                    Therefore, $[\sigma]_{\tabisim} \in V$ (either it is added in line~\ref{alg:absts:add-v} or it was already in $V$ before) and $([\sigma^*]_{\tabisim}, [\sigma]_{\tabisim})$ is added to $E$ in line~\ref{alg:absts:add-e}.
            \end{enumerate}
        \end{description}
        \textbf{$\Leftarrow$:}
        Let $p = [S_0]_{\tabisim} \rightarrow^* [\sigma]_{\tabisim}$ be a path in $T$.
        We show by induction on the length $n = |p|$ of $p$ that
        (1) $\bat \models \exec(\sigma)$ and (2) $p$ is a path in $\absts$.
        \begin{description}
          \item[Base case.]
            Let $n = 0$, i.e., $p$ is the empty path. Clearly, the empty path is also a path in \absts.
            Also, by definition, $\bat \models \exec(S_0)$.
          \item[Induction step.]
            Let $p = [S_0]_{\tabisim} \rightarrow^* [\sigma^*]_{\tabisim} \rightarrow [\sigma]_{\tabisim}$.
            By induction, $\bat \models \exec(\sigma^*)$ and $[S_0]_{\tabisim} \rightarrow^* [\sigma^*]_{\tabisim}$ is a path in \absts.
            It remains to be shown that $\bat \models \exec(\sigma)$ and $[\sigma^*]_{\tabisim} \rightarrow [\sigma]_{\tabisim}$ is a transition in \absts.
            Clearly, $\sigma = \doop(\alpha, \sigma^*)$ for some ground action term $\alpha$.
            From the definition of \autoref{alg:regiontransitionsystem}, it directly follows that $\bat \models \poss(\alpha, \sigma^*)$ and so $\bat \models \exec(\sigma)$.
            As $\bat \models \exec(\sigma)$, by definition of \absts, $[\sigma^*]_{\tabisim} \rightarrow [\sigma]_{\tabisim}$ is a transition in \absts.
        \end{description}
    We now show that the algorithm terminates.
    By \autoref{thm:regression-decidable} checking whether $\bat_0 \cup \bat_c \models \reg{\poss(\alpha, \sigma)}$ is decidable.
    Also, by \autoref{lma:sitRelatableIsDecidable}, checking for each new situation $\doop(\alpha, \sigma)$ whether there is an existing state $[\sigma]_{\tabisim} \in V$ (line~\ref{alg:abs:check-v}) is decidable.
    Furthermore, by \autoref{lma:absts-finite}, \absts is finite, and so the loop eventually terminates.
\end{proofE}
\begin{theoremE}[][normal]\label{thm:main-result}
    Reachability for clocked \acp{BAT} is decidable.
\end{theoremE}
\begin{proofE}
  By \autoref{lma:absts-reachability}, it is sufficient to consider the time-abstract transition system \absts.
  By \autoref{lma:alg-returns-absts}, \autoref{alg:regiontransitionsystem} terminates and computes \absts.
  We now show how to use \absts for deciding the reachability problem of some property $\phi$ with maximal constant $\leq \maxconst$: 
  By construction, every state of \absts is an equivalence class wrt.\ \tabisim of executable situations, of which there are finitely many.
  By \autoref{lma:bisim:static-equivalence}, it is sufficient to consider one situation $\sigma^* \in [\sigma]_{\tabisim}$ for each such state.
  Furthermore, by \autoref{thm:regression}, we can check whether $\bat \models \phi[\sigma^*]$ by regression.
  If such a state exists, return ``reachable'', otherwise return ``not reachable''.
\end{proofE}

\begin{example}
    We can check the following reachability queries:
    \begin{align*}
        \phi_1 &\eqdef \forall o. \hasCoffee(o)
        \\
        \phi_2 &\eqdef \forall o. \rfluent{strong}(o) \equiv \rfluent{wantStrong}(o)
        \\
        \phi_3 &\eqdef c_{\action{glob}}() \leq 1 \land \rfluent{strong}(\mi{Mug}_1)
    \end{align*}
    A situation satisfying $\phi_1$ is reachable, e.g., with 
    \begin{multline*}
      \sigma_1 = \doop([\sac{brew}(\mi{Pot});\action{wait}(1);\allowbreak\eac{brew}(\mi{Pot});\allowbreak\sac{pour}(\mi{Pot}, \mi{Mug}_1);
      \eac{pour}(\mi{Pot}, \mi{Mug}_1);
      \\
      \sac{pour}(\mi{Pot}, \mi{Mug}_2);\allowbreak\eac{pour}(\mi{Pot}, \mi{Mug}_2)], S_0)
    \end{multline*}
    For $\phi_2$, a different action sequence is required, as the pot may not be filled with strong and weak coffee at the same time.
    Instead, we brew coffee directly in a mug:
    \begin{multline*}
      \sigma_2= \doop([\sac{brew}(\mi{Pot});\allowbreak \sac{brew}(\mi{Mug}_1);\allowbreak \action{wait}(1);\allowbreak \eac{brew}(\mi{Pot});\allowbreak \action{wait}(2);
      \eac{brew}(\mi{Mug}_1);
      \\
      \sac{pour}(\mi{Pot}, \mi{Mug}_2);\allowbreak \action{wait}(1); \allowbreak \eac{pour}(\mi{Pot}, \mi{Mug}_2)],\allowbreak S_0)
    \end{multline*}
    Finally, there is no reachable situation that satisfies $\phi_3$, as it would require providing strong coffee after one time unit, while it takes at least two time units to prepare strong coffee.
\end{example}

\section{Realizing \golog Programs}\label{sec:program-realization}

In this section, we show how we can utilize reachability for clocked \acp{BAT} to decide realizability for a \golog program $\Delta$.
The goal is to check whether the non-determinisms in $\Delta$ can be resolved and action timepoints can be determined such that the resulting action sequence is executable and results in a final configuration of the program.
Similar to the above, we create a transition system $\absts[\Delta]$ from a given program $\Delta$ and check whether a final program state is reachable in $\absts[\Delta]$.

\subsection{Golog Transition Semantics}
We consider programs in \congolog~\cite{de_giacomo_congolog_2000} based on clocked \acp{BAT}. 
A \emph{program} is a pair $\Delta = (\bat, \delta)$, where \bat is \iac{BAT} and $\delta$ is a program expression built from the following grammar:
\[
  \delta ::= \alpha \smid \phi? \smid \delta; \delta \smid \delta \vert \delta \smid  \pi x.\, \delta \smid \delta^* \smid \delta\|\delta
\]
Program expressions consist of actions $\alpha$, tests $\phi?$, sequence of actions $\delta_1; \delta_2$, non-deterministic branching $\delta_1 \vert \delta_2$, non-deterministic choice of argument $\pi x.\, \delta$, non-deterministic iteration $\delta^*$, and concurrent execution $\delta_1 \| \delta_2$.
We require that each action term is time-suppressed (i.e., $\action{wait}()$ has no argument) and each test $\phi?$ tests a situation-suppressed clocked formula $\phi$.
We write $\nil \eqdef \true?$ for the empty program that always succeeds.
As usual, conditionals and loops can be defined as macros, i.e., 
$\gif \phi \gdo \delta_1 \gelse \delta_2 \gfi \eqdef [\phi?; \delta_1] \vert [\neg \phi?; \delta_2]$
and
$\gwhile \phi \gdo \delta \gdone \eqdef [\phi?; \delta]^*; \neg \phi?$.
We call $\Delta = (\bat, \delta)$ a \emph{clocked program} if \bat is a clocked \ac{BAT} and
we assume that $\maxconst \in \naturals$ is the maximal constant mentioned in \bat and $\delta$.
We adapt the \golog transition semantics from \congolog such that each transition step follows a single primitive action, corresponding to \emph{synchronized conditionals} known from \indigolog~\cite{de_giacomo_indigolog_2009}.
We also extend the relation $\gtrans$ to include the action such that $\gtrans(\delta, s, a, \delta', s')$ is true if there is a single-step transition with action $a$ in the program $\delta$ starting in situation $s$ and resulting in situation $s'$ with remaining program $\delta'$.
Finally, we assume that the program $\delta$ does not fix the time of occurrence for \emph{wait} and instead the time is determined in the transition semantics.
Therefore, \action{wait} actions in $\delta$ do not have a time argument $t$ and the timepoint is instead fixed in the transition rule for a \action{wait} action.\footnote{We may still enforce a certain time delay by using clock formulas in tests $\phi?$, as demonstrated in \autoref{ex:program-realization}.}
We also write $\gtrans^*(\delta, s, \vec{a}, \delta', s')$ for the transitive closure of $\gtrans$ such that $\vec{a}$ is the sequence of actions resulting from following the transitions according to $\gtrans$:
\begin{align*}
    \gtrans(a, s, a, \delta', s') \equivspace
    {}&\poss(a, s) \wedge \delta' = \nil \wedge s' = \doop(a, s)
    \\
    \gtrans(\action{wait}(), s, a, \delta', s') \equivspace
    {}&\exists t.\, a = \action{wait}(t) \wedge \delta' = \nil \wedge s' = \doop(a, s)
    \\
    \gtrans(\phi?, s, a, \delta', s') \equivspace {}&\false
    \\
    \gtrans(\delta_1; \delta_2, s, a, \delta', s') \equivspace
    {}&\exists \gamma.\, \delta' = (\gamma; \delta_2) \wedge \gtrans(\delta_1, s, a, \gamma, s')
    \\
    &\quad \vee \gfinal(\delta_1, s) \wedge \gtrans(\delta_2, s, a, \delta', s')
    \\
    \gtrans(\delta_1 \vert \delta_2, s, a, \delta', s') \equivspace
    {}&\gtrans(\delta_1, s, a, \delta', s')
    \vee \gtrans(\delta_2, s, a, \delta', s')
    \\
    \gtrans(\pi x. \delta, s, a, \delta', s') \equivspace
    {}&\exists v.\, \gtrans(\delta^x_v, s, a, \delta', s')
    \\
    \gtrans(\delta^*, s, a, \delta', s') \equivspace 
    {}&\exists \gamma.\, (\delta' = \gamma; \delta^*)
    \wedge \gtrans(\delta, s, a, \gamma, s')
    \\
    \gtrans(\delta_1 \| \delta_2, s, a, \delta', s') \equivspace
    {}&\exists \gamma.\, \delta' = (\gamma \| \delta_2) \wedge \gtrans(\delta_1, s, a, \gamma, s')
      \\
      &\quad \vee \exists \gamma.\, \delta' = (\delta_1 \| \gamma) \wedge \gtrans(\delta_2, s, a, \gamma, s')
\end{align*}
$\gfinal(\delta, s)$ is true if program $\delta$ may terminate in situation $s$:
\begin{align*}
      \gfinal(a, s) \equivspace {}& \false
      &\gfinal(\phi?,  s) \equivspace {}& \phi[s]
      \\
      \gfinal(\delta_1; \delta_2, s) \equivspace {}& \gfinal(\delta_1, s) \wedge \gfinal(\delta_2, s)
      \\
      \gfinal(\delta_1 \vert \delta_2, s) \equivspace {}& \gfinal(\delta_1, s) \vee \gfinal(\delta_2, s)
      \\
      \gfinal(\pi x. \delta, s) \equivspace {}& \exists v.\, \gfinal(\delta^x_v, s)
      &
      \gfinal(\delta^*, s) \equivspace {}& \true
      \\
      \gfinal(\delta_1 \| \delta_2, s) \equivspace {}& \gfinal(\delta_1, s) \wedge \gfinal(\delta_2, s)
\end{align*}

\begin{example}\label{ex:program-realization}
    We consider the following program $\delta$:
    \begin{sizeddisplay}
    \begin{align*}
    \big( &\sac{brew}(\mi{Pot}); \exists m (\rfluent{wantStrong}(m)) \equiv c_{\action{brew}}(\mi{Pot}) \geq 2?;\eac{brew}(\mi{Pot}); \\
    &\gwhile(\exists m \neg \hasCoffee(m)) \gdo
     \pi m.\,
    \sac{pour}(\mi{Pot}, m); \eac{pour}(\mi{Pot}, m);
    \gdone \big) \;\|\; \action{wait}()^*
    \end{align*}
    \end{sizeddisplay}
    The program first brews coffee.
    If there is a mug that should be filled with strong coffee, it waits until $c_{\action{brew}}(\mi{Pot}) \geq 2$, otherwise it stops brewing earlier.
    Afterwards, it fills all mugs.
\end{example}

\subsection{Program Realization}\label{ss:program-realization}

\begin{algorithm}[tb]
\caption{Construction of the time-abstract transition system \absts[\Delta].}
\label{alg:regionprogramtransitionsystem}
\begin{algorithmic}[0]
\smallfont
\State $\mi{Open} \gets \{ (S_0, \delta_0) \}$;
$\mi{V} \gets \emptyset$;
$\mi{E} \gets \emptyset$
\While{$\mi{Open} \neq \emptyset$}
    \State $(\sigma, \delta) \gets \Call{pop}{\mi{Open}}$
    \State $\mi{Acts} \gets \{ A(\vec{\rho}) \mid A \in \actions \setminus \{ \action{wait} \}, \vec{\rho} \in \mathcal{O}^{|A|} \} \cup \{ \action{wait}(\tau) \mid \tau \in \Call{TSuccs}{\sigma, \maxconst} \}$
    \ForAll{$\alpha \in \mi{Acts}$}
        \If{$\bat \models \gtrans(\delta, \sigma, \alpha, \delta', \sigma')$}
            \If{$([\sigma']_{\tabisim}, \delta') \not\in \mi{V}$}
                \State $\mi{V} \gets \mi{V} \cup \{ ([\sigma']_{\tabisim}, \delta') \}$
                \State $\mi{Open} \gets \mi{Open} \cup \{ (\sigma', \delta') \}$
            \EndIf
            \State $\mi{E} \gets \mi{E} \cup \{ (([\sigma]_{\tabisim}, \delta), ([\sigma']_{\tabisim}, \delta')) \}$
        \EndIf
    \EndFor
\EndWhile
\State \Return $(\mi{V},([S_0]_{\tabisim}, \delta_0),\mi{E})$
\end{algorithmic}
\end{algorithm}

We describe how to construct a transition system $\absts[\Delta]$ to determine a realization of a given program $\Delta$  similar to \absts from above.
We first formally define program realization:
\begin{definition}[Program Realization]
Let \bat be a \ac{BAT} and $\delta$ a program. An action sequence $\vec{\alpha} = \la \alpha_1, \ldots, \alpha_n \ra$ is a realization of $\delta$ wrt.\ \bat if $\bat \models \gtrans^*(\delta, S_0, \vec{\alpha}, \delta', \sigma') \wedge \gfinal(\delta', \sigma')$.
The corresponding decision problem is to decide
whether such a realization exists.
\end{definition}

First, we can show that the equivalence relation \tabisim can also be applied to the transition semantics, i.e., bisimilar situations cannot be distinguished by possible program transitions:
\begin{lemmaE}\label{lma:trans-equivalence}
    Let $\sigma_1, \sigma_2$ be two situations with $\sigma_1 \tabisim \sigma_2$ and $\bat \models \gtrans(\delta, \sigma_1, \alpha, \delta^*, \sigma_1')$.
    \begin{enumerate}
        \item If $\alpha = \action{wait}(\tau_1)$ for some $\tau_1 \in \realpos$, then there is $\tau_2 \in \textsc{TSuccs}(\sigma_2, K)$ such that $\gtrans(\delta, \sigma_2, \action{wait}(\tau_2), \delta^*, \sigma_2')$ and $\sigma_1' \tabisim \sigma_2'$.
        \item If $\alpha = A(\vec{\rho})$ for some action type $A \neq \action{wait}$, then $\gtrans(\delta, \sigma_2, \alpha, \delta^*, \sigma_2')$ and $\sigma_1' \tabisim \sigma_2'$.
    \end{enumerate}
\end{lemmaE}
\begin{proofE}
    First, note that by \autoref{lma:bisim:static-equivalence}, for every clocked formula $\phi$ uniform in $s$, $\bat \models \phi[\sigma_1]$ iff $\bat \models \phi[\sigma_2]$ and so it directly follows that $\bat \models \gfinal(\delta, \sigma_1)$ iff $\bat \models \gfinal(\delta, \sigma_2)$.
    \begin{enumerate}
        \item Let $\alpha = \action{wait}(\tau_1)$ for some $\tau_1 \in \realpos$.
            By \autoref{lma:tsucc}, there is a $\tau_2 \in \textsc{TSuccs}(\sigma_2, K)$ such that $\clockval_{\sigma_1} + \tau_1 \regequiv \clockval_{\sigma_2} + \tau_2$.
            Let $\sigma_2' = \doop(\action{wait}(\tau_2), \sigma_2)$.
            With $\bat \models \poss(\action{wait}(t), s) \equiv \top$, it follows that $\gtrans(\delta, \sigma_2, \action{wait}(\tau_2), \delta^*, \sigma_2')$.
            It remains to be shown that $\sigma_1' \tabisim \sigma_2'$.
            Analogously to \autoref{thm:equivRelation}:
            \begin{enumerate}
                \item Let $R \in \fluents$ be a relational fluent and $\pvec{\rho}'$ a ground tuple of arguments.
                We show that $\bat \models R(\pvec{\rho}', \sigma_1')$ iff $\bat \models R(\pvec{\rho}', \sigma_2')$.
                There is \iac{SSA} for $R$ of the form $R(\vec{o}, \doop(a, s)) \equiv \phi_R(\vec{o}, a, s)$, where $\phi_R(\vec{o}, a, s)$ is a clocked formula uniform in $s$.
                As $\sigma_1 \tabisim \sigma_2$, it directly follows from \autoref{lma:bisim:static-equivalence} that $\bat \models \phi_R(\pvec{\rho}', \action{wait}(\tau_1), \sigma_1)$ iff $\bat \models \phi_R(\pvec{\rho}', \action{wait}(\tau_2), \sigma_2)$ and therefore $\bat \models R(\pvec{\rho}', \sigma_1')$ iff $\bat \models R(\pvec{\rho}', \sigma_2')$.
                \item
                Let $c$ be a func.\ fluent and $\omega = c(\pvec{\rho}')$.
                By definition of the \ac{SSA} for  $c$, it follows that $\clockval_{\sigma_1'}(\omega) = \clockval_{\sigma_1}(\omega) + \tau_1$ and $\clockval_{\sigma_2'}(\omega) = \clockval_{\sigma_2}(\omega) + \tau_2$.
                Thus, we can set $f'(\omega, \clockval_{\sigma_1'}(\omega)) = (\omega, \clockval_{\sigma_2'}(\omega))$.
                As $\clockval_{\sigma_1} + \tau_1 \regequiv \clockval_{\sigma_2} + \tau_2$, it follows that $f'$ is a bijection witnessing \tabisim.
            \end{enumerate}
        \item Let $\alpha = A(\vec{\rho})$ for some action type $A \neq \action{wait}$ and so $\sigma_1' = \doop(\alpha, \sigma_1)$.
            There is a precondition axiom of the form $\poss(A(\vec{o}), s) \equiv \phi_A(\vec{o}, s)$ where $\phi_A(\vec{o}, s)$ is a clocked formula uniform in $s$.
            As $\gtrans(\delta, \sigma_1, \alpha, \delta^*, \sigma_1')$, it is clear that $\bat \models \phi_A(\vec{\rho}, \sigma_1)$ and with $\sigma_1 \tabisim \sigma_2$, also $\bat \models \phi_A(\vec{\rho}, \sigma_2)$.
            Therefore, $\bat \models \poss(A(\vec{\rho}), \sigma_2)$ and so $\bat \models \gtrans(\delta, \sigma_2, \alpha, \delta^*, \sigma_2')$ with $\sigma_2' = \doop(\alpha, \sigma_2)$.
            It remains to be shown that $\sigma_1' \tabisim \sigma_2'$.
            Analogously to \autoref{thm:equivRelation}:
            \begin{enumerate}
                \item Let $R \in \fluents$ be a relational fluent and $\pvec{\rho}'$ a ground tuple of arguments.
                We show that $\bat \models R(\pvec{\rho}', \sigma_1')$ iff $\bat \models R(\pvec{\rho}', \sigma_2')$.
                There is \iac{SSA} for $R$ of the form $R(\vec{o}, \doop(a, s)) \equiv \phi_R(\vec{o}, a, s)$, where $\phi_R(\vec{o}, a, s)$ is a clocked formula uniform in $s$.
                As $\sigma_1 \tabisim \sigma_2$, it directly follows from \autoref{lma:bisim:static-equivalence} that $\bat \models \phi_R(\pvec{\rho}', A(\vec{\rho}), \sigma_1)$ iff $\bat \models \phi_R(\pvec{\rho}', A(\vec{\rho}), \sigma_2)$ and therefore $\bat \models R(\pvec{\rho}', \sigma_1')$ iff $\bat \models R(\pvec{\rho}', \sigma_2')$.
                \item We construct a bijection $f': \clockval_{\sigma_1'} \rightarrow \clockval_{\sigma_2'}$ witnessing \tabisim.
                Let $c$ be a functional fluent and $\omega = c(\pvec{\rho}')$.
                There is \iac{SSA} of the form $
                  c(\vec{o}, \doop(a, s)) = y \equivspace
                  \exists t.\, a = \action{wait}(t) \wedge y = c(\vec{o}, s) + t
                  \vee \neg \exists t.a = \action{wait}(t) \wedge (\phi_c(\vec{o}, a, s) \wedge y = 0 
                  \vee \neg \phi_c(\vec{o}, a, s) \wedge y = c(\vec{o}, s))$.
                Thus, $\omega[\sigma_1'] = 0$ if $\bat \models \phi_c(\vec{\rho}, \alpha, \sigma_1)$ and $\omega[\sigma_1'] = \omega[\sigma_1]$ otherwise, similarly for $\omega[\sigma_2']$.
                Now, as $\phi_c(\vec{\rho}, a, s)$ is a clocked formula uniform in $s$, by \autoref{lma:bisim:static-equivalence}, $\bat \models \phi_c(\vec{\rho}, \alpha, \sigma_1)$ iff $\bat \models \phi_c(\vec{\rho}, \alpha, \sigma_2)$.
                Hence, for each $\omega$, we can set $f'(\omega, 0) = (\omega, 0)$ if $\bat \models \phi_c(\vec{\rho}, \alpha, \sigma_1)$ and $f'(\omega, \clockval_{\sigma_1}(\omega)) = (\omega, \clockval_{\sigma_2}(\omega))$ otherwise.
                \qedhere
            \end{enumerate}
    \end{enumerate}
\end{proofE}

This allows us to use time-abstract bisimulations for determining realizability.
We adapt \autoref{alg:regiontransitionsystem} such that states also contain the remaining program.
The resulting algorithm is shown in \autoref{alg:regionprogramtransitionsystem}.
The algorithm is correct, i.e., the transition system contains a final program configuration iff a realization exists:
\begin{lemmaE}\label{lma:ts-realizability}
    Let $\Delta = (\bat, \delta)$ be a program over a clocked \ac{BAT} \bat
    and $\absts[\Delta]$ the transition system returned by \autoref{alg:regionprogramtransitionsystem}.
    Then $\absts[\Delta]$ contains a state $([\sigma']_{\tabisim}, \delta')$ reachable from $([S_0]_{\tabisim}, \delta)$ with $\bat \models \gfinal(\delta', \sigma')$ iff there is a realization of program $\Delta$.
\end{lemmaE}
\begin{proofE}
    We first show that $([\sigma]_{\tabisim}, \delta')$ is reachable in $\absts[\Delta]$ iff $\bat \models \gtrans^*(\delta, S_0, \vec{\alpha}, \delta', \sigma)$.
    ~ \\
    \textbf{$\Rightarrow$:}
    Let $p = ([S_0]_{\tabisim}, \delta) \rightarrow^* ([\sigma']_{\tabisim}, \delta')$ be a path in $\absts[\Delta]$.
    We show by induction on the length $n$ of $p$ that
    $\bat \models \gtrans^*(\delta, S_0, \vec{\alpha}, \delta', \sigma')$ for some $\vec{\alpha}$ with $|\vec{\alpha}| = n$:
    \begin{description}
      \item[Base case.]
        Let $n = 0$ and therefore $\sigma' = S_0$ and $\delta' = \delta$.
        Clearly, $\bat \models \gtrans^*(\delta, S_0, \la\ra, \delta, S_0)$.
      \item[Induction step.]
        Let $p = ([S_0]_{\tabisim}, \delta) \rightarrow^* ([\sigma^*]_{\tabisim}, \delta^*) \rightarrow ([\sigma']_{\tabisim}, \delta')$.
        By induction, $\bat \models \gtrans^*(\delta, S_0, \vec{\alpha}, \delta^*, \sigma^*)$ for some $\vec{\alpha}$ with $|\vec{\alpha}| = n$.
        By construction of $\absts[\Delta]$, $\bat \models \gtrans(\delta^*, \sigma^*, \alpha, \delta', \sigma')$ for some $\alpha$ and therefore also $\bat \models \gtrans^*(\delta, S_0, \vec{\alpha} \cdot \alpha, \delta', \sigma')$.
    \end{description}
    \textbf{$\Leftarrow$:}
    Let $\bat \models \gtrans^*(\delta, S_0, \vec{\alpha}, \delta', \sigma')$.
    We show by induction on the length $n$ of $\vec{\alpha}$ that there is a path $p = ([S_0]_{\tabisim}, \delta) \rightarrow^* ([\sigma']_{\tabisim}, \delta')$ in $\absts[\Delta]$ with length $n$.
    \begin{description}
      \item[Base case.]
        Let $n = 0$ and therefore $\sigma' = S_0$ and $\delta' = \delta$.
        Clearly, there is an empty path in $\absts[\Delta]$ starting in $([S_0]_{\tabisim}, \delta)$.
      \item[Induction step.]
        Let $\bat \models \gtrans^*(\delta, S_0, \vec{\alpha}, \delta^*, \sigma^*)$ for some $\vec{\alpha}$ with $|\vec{\alpha}| = n$ and let $\bat \models \gtrans(\delta^*, \sigma^*, \alpha, \delta', \sigma')$.
        By induction, there is a path $p = ([S_0]_{\tabisim}, \delta) \rightarrow^* ([\sigma^*]_{\tabisim}, \delta^*)$ with length $n$.
        By \autoref{lma:trans-equivalence}, there is a $\alpha'$ such that $\bat \models \gtrans(\delta^*, \sigma^*, \alpha', \delta', \sigma'')$ and $\sigma'' \tabisim \sigma'$.
        Hence, $([\sigma^*]_{\tabisim}, \delta^*) \rightarrow  ([\sigma']_{\tabisim}, \delta')$ is a transition in \absts[\Delta].
    \end{description}
    Now, as $([\sigma]_{\tabisim}, \delta')$ is reachable in $\absts[\Delta]$ iff $\bat \models \gtrans^*(\delta, S_0, \vec{\alpha}, \delta', \sigma)$ for some $\vec{\alpha}$, it follows with \autoref{lma:bisim:static-equivalence} that $\bat \models \gfinal(\delta', \sigma)$ iff $\vec{\alpha}$ is a realization of $\Delta$.
\end{proofE}
Moreover, the transition system is finite:
\begin{lemmaE}[][normal]\label{lma:program-ts-finite}
    $\absts[\Delta]$ has finitely many states.
\end{lemmaE}
\begin{proofE}
    As \objects and \actions are finite, each state only has finitely many successors.
    Furthermore, as there are only finitely many fluents \fluents and objects \objects, \tabisim has finitely many equivalence classes.
    Finally, there are finitely many reachable sub-programs from $\delta$ \cite{classenLogicNonterminatingGolog2008}.
    Therefore, the time-abstract transition system \absts[\Delta] has finitely many states.
\end{proofE}
Combining the results, we obtain:
\begin{theoremE}[][normal]
    The realization problem for a clocked program $\Delta = (\bat, \delta)$ is decidable.
\end{theoremE}
\begin{proofE}
    By \autoref{lma:ts-realizability}, it is sufficient to check whether there is a reachable state in \absts[\Delta] that corresponds to a final program  configuration.
    By \autoref{lma:program-ts-finite}, \absts[\Delta] is finite and so explicitly constructing it is feasible.
    Finally, checking whether $\bat \models \gtrans(\delta, \sigma, \alpha, \delta', \sigma')$ reduces to checking entailment of clocked formulas uniform in $\sigma$ (in particular, checking $\bat \models \poss(\alpha, \sigma)$ and $\bat \models \phi[\sigma]$ for tests $\phi?$).
    By \autoref{thm:regression-decidable}, this is decidable.
\end{proofE}

We conclude by presenting a realization of the example:
\begin{example}
    The following sequence is a realization of the program from \autoref{ex:program-realization}:
\begin{sizeddisplay}
\begin{align*}
&\sac{Brew}(\mi{Pot}); \action{wait}(2); \eac{Brew}(\mi{Pot}); \sac{Pour}(\mi{Pot},\mi{Mug}_1);\\
&\eac{Pour}(\mi{Pot},\mi{Mug}_1); \sac{Pour}(\mi{Pot},\mi{Mug}_2); \eac{Pour}(\mi{Pot},\mi{Mug}_2);
\end{align*}    
\end{sizeddisplay}

However, the program does not fill every mug with the correct kind of coffee.
In fact, let $\delta' \eqdef \delta ; \forall m (\isStrong(m) \equiv \rfluent{wantStrong}(m))?$ be the program $\delta$ augmented with a check whether every mug was filled with the right kind of coffee.
Then, no realization of $\delta'$  exists:
As the program brews coffee only once, it fills all mugs with the same coffee.
\end{example}

\section{Conclusion}\label{sec:conclusion}
We investigated how continuous time can be modeled in the \acf{SC},
focusing on \emph{reachability}, which asks whether there is a reachable situation that satisfies some  given formula $\phi$, and \emph{program realization}, the task to determine an action sequence corresponding to a successful execution of a given \golog program.
For these problems, the standard approaches for modeling time in the \ac{SC} are unsuitable, because reachability (and thus also realizability) is undecidable.
We presented an alternative approach based on \emph{clocks}, i.e., real-valued fluents with restricted \aclp{SSA}, and described sound and complete algorithms that explicitly construct a finite transition system, showing that both problems are decidable.

Our approach is restricted to \aclp{BAT} with a finite set of objects.
Since quantifiers can be eliminated by grounding, our setting is essentially that of the propositional situation calculus \cite{degiacomoReactiveSynthesisGolog2026}, modulo the real-valued time of \action{wait}.
For future work, it may be interesting to extend this approach towards more expressive formalisms that retain decidability, e.g., \emph{bounded action theories}~\cite{de_giacomo_bounded_2016},
or the two-variable fragment of the \ac{SC}~\cite{gu_description_2010}, which has been used for verifying \golog programs over local-effect~\cite{clasen_exploring_2014} and acyclic~\cite{zarries_decidable_2016} \acp{BAT}, and for synthesizing \golog policies against uncontrollable environments with LTL\textsubscript{f} goals \cite{hofmannLTLfSynthesisFirstorder2025}.
Extending our notion of time to the synthesis setting is a promising direction.

\appendix

\bibliography{zotero}

\ifthenelse{\boolean{showappendix}}{
\clearpage

\ifextendedappendix
\section{Appendix}
In this section, we provide complete definitions for \acp{TA} and \acfp{2CM}.

\subsection{Timed Automata}

\acp{TA} extend classical transition systems with time by adding a finite number of clocks that allow measuring time.
Clocks increase their valuations \emph{continuously} while the system is in a certain location (mode) and may be reset to zero when switching locations via possibly guarded transitions.
Guard conditions are specified as clock constraints.

\begin{definition}[Clock Constraint]
  Let $\taclocks$ be a set of clocks.
  The set $\clockconstraints(\taclocks)$ of \emph{clock constraints} $\clockconstraint$ is defined by the grammar: $\clockconstraint ::= x \bowtie c \smid \clockconstraint_1 \wedge \clockconstraint_2$ where $\mathop{\bowtie}\in\{<,\leq,=,\geq,>\}$, $x \in \taclocks$ is a clock, and $c \in \naturals$ is a constant.
\end{definition}

\begin{definition}[Timed Automata] A \acl{TA} (c.f.~\cite{alur_timed_1999}) $\ta = (\talocs, \taloc^0, \talocs^F, \taalph, \taclocks, \taswitches)$ is a tuple with
a finite set of locations $\talocs$;
an initial location $\taloc^0$;
a set of final locations $\talocs^F \subseteq  L$;
a finite set of labels $\taalph$;
a finite set of clocks $\taclocks$;
a finite set of \emph{switches} $E \subseteq \talocs \times \taalph \times \clockconstraints(\taclocks) \times 2^\taclocks \times \talocs$ which defines possible switches $(l,k,g,Y,l')$ from location $l$ to location $l'$, labelled with $k$, only possible if the guard condition $g$ is satisfied. All clocks in $Y$ are reset to zero upon taking the switch.
\end{definition}

The behavior of a \ac{TA} can be described by means of a transition system $\talts$ where a state combines a location and a clock valuation.
The transition relation between states of $\talts$ combines a continuous step (letting a specific amount of time pass) and a discrete switch of locations.

\begin{definition}\label{def:ta-lts}
  Let $\ta = (L, l_0, L_F, \Sigma, X, E)$ be a \ac{TA}.
  The corresponding \acf{LTS} $\talts = \left(\taltsstates, \taltsstate^0, \taltsstates^F, \taltsalph, \taltstrans{}{}\right)$ is defined as follows:
  \begin{itemize}
    \item A state $\taltsstate \in \taltsstates$ of $\talts$ is a pair $\left(\taloc, \clockval\right)$ such that
      \begin{enumerate}
        \item \taloc{} is a location of \ta,
        \item \clockval is a clock valuation for the clocks \taclocks of \ta
      \end{enumerate}
    \item The initial state $(\taloc^0, \clockval_0)$ consists of the initial location $l_0$ and a clock valuation $\clockval_0$ where all clocks are zero-initialized,
    \item The final states $\taltsstates^F = \{ (\taloc, \clockval) \mid \taloc \in \talocs^F \}$ are those states that contain a final location of \ta,
    \item The labels $\taltsalph$ consist of the labels of the \ac{TA} and time increments, i.e., $\taltsalph = \taalph \times \mathbb{R}$.
    \item A transition $\left(\taloc, \clockval\right) \taltstrans{\delta}{a} \left(\taloc', \clockval'\right)$ consists of two steps:
      \begin{enumerate}
        \item \emph{Elapse of time:} All clocks are incremented by some time increment $\delta \in \realpos$
        \item \emph{Switch of location:} The location changes based on a switch $\left(\taloc, a, \varphi, Y, \taloc'\right) \in \taswitches$, where the clock constraint $\varphi$ must be satisfied by the incremented clocks $\clockvaluation^*$ and $Y$ specifies which clocks are reset after the transition, i.e., $\clockvaluation^* \models \varphi$ and $\clockvaluation' = \clockvaluation^*[Y \eqdef 0]$.
      \end{enumerate}
  \end{itemize}
\end{definition}

A \emph{path} from $\ltsstate$ to $\ltsstate'$ is a finite sequence of transitions $\ltsstate \ltstrans \ltsstate^1 \ltstrans \ldots \ltstrans \ltsstate'$.
A \emph{run} on \ta is a path starting in the initial state $\ltsstate^0$.
We denote the set of all finite runs of \ta with $\finruns(\ta)$.
A finite run is \emph{accepting} if it ends in an accepting state $\ltsstate \in \taltsstates^F$.

\begin{definition}[Language of a \ac{TA}]\label{def:ta-language}
  Given a (finite) run
  $p = \left(\tastate_0, \clockvaluation_0\right) \taltstrans{\delta_1}{a_1} \allowbreak \cdots \taltstrans{\delta_n}{a_n} \allowbreak \left(\tastate_{n+1}, \clockvaluation_{n+1}\right) \cdots \allowbreak \in \finruns(\ta)$ on a \ac{TA} \ta.
  The \emph{timed word induced by $p$} is the timed word $\tw(p) = \left(a_1, \delta_1\right) \allowbreak \left(a_2, \delta_1 + \delta_2\right) \cdots (a_n, \sum_1^n \delta_i) \cdots$.
  The \emph{language of finite words of \ta} is the set $\lang^*(\ta) = \{ \tw(p) \mid p \in \accfinruns(\ta) \}$.
\end{definition}

\subsection{2-Counter Machines}

We use \acfp{2CM} as defined by~\cite{bouyer_updatable_2004}:
\begin{definition}[2-Counter Machines]
  A \acf{2CM} is a finite set of labeled instructions over two counters $c_1$ and $c_2$.
  There are two types of instructions:
  \begin{enumerate}
    \item An \emph{incrementation instruction} of counter $x \in \{ c_1, c_2 \}$:
      \[
        p:\: x \eqdef x + 1; \;\mathbf{goto}\; q
      \]
      The instruction increments counter $x$ by one and then goes to the next instruction $q$.
    \item A \emph{decrementation instruction} of counter $x \in \{ c_1, c_2 \}$:
      \[
        p:\: \textbf{if}\; x > 0
        \begin{cases}
          \mathbf{then}\; x \eqdef x - 1; \; \mathbf{goto}\; q
          \\
          \mathbf{else}\; \mathbf{goto}\; r
        \end{cases}
      \]
      The instruction branches on $x$: If $x$ is larger than $0$, then it decrements $x$ and goes to instruction $q$.
      Otherwise, it does not change $x$ and directly goes to instruction $r$.
  \end{enumerate}
  The machine starts with instruction $s_0$ and with  counter values $c_1 = c_2 = 0$ and stops at a special instruction $\textbf{HALT}$.
  The \emph{halting problem} for a \ac{2CM} is to decide whether a machine reaches the instruction $\textbf{HALT}$.
\end{definition}

\begin{theorem}
  The halting problem for \acp{2CM} is undecidable~\cite{minsky_computation_1967}.
\end{theorem}

\fi
}{}

\section{Proofs}\label{sec:proofs}
\printProofs
\end{document}